\documentclass{article}
\usepackage{amssymb}
\usepackage{amsmath}
\usepackage{graphicx}
\usepackage{physics}
\usepackage{makecell}
\usepackage{caption}
\usepackage{algorithm}
\usepackage{algpseudocode}
\usepackage[svgnames,x11names,table, dvipsnames]{xcolor}

\definecolor{Burgundy}{RGB}{144,0,32}
\definecolor{Burgundy1}{RGB}{128,0,32}
\definecolor{Burgundy2}{RGB}{158,5,8}
\definecolor{VividBurgundy}{RGB}{159,29,53}

\usepackage{amsthm}
\usepackage{wrapfig}
\usepackage[preprint]{corl_2025} 

\newtheorem{ass}{\textbf{Assumption}}
\newtheorem{pro}{\textbf{Proposition}}[section]

\newtheorem{rem}{\textbf{Remark}}

\title{Rapid Mismatch Estimation via Neural Network Informed Variational Inference}

\author{
  Mateusz Jaszczuk \quad Nadia Figueroa\\
  GRASP Lab, University of Pennsylvania\\
  \texttt{\{jaszczuk, nadiafig\}@seas.upenn.edu} \\
}

\begin{document}
\maketitle
\begin{abstract} 
With robots increasingly operating in human-centric environments, ensuring soft and safe physical interactions, whether with humans, surroundings, or other machines, is essential. While compliant hardware can facilitate such interactions, this work focuses on impedance controllers that allow torque-controlled robots to safely and passively respond to contact while accurately executing tasks. From inverse dynamics to quadratic programming-based controllers, the effectiveness of these methods relies on accurate dynamics models of the robot and the object it manipulates. Any model mismatch results in task failures and unsafe behaviors. Thus, we introduce Rapid Mismatch Estimation (RME), an adaptive, controller-agnostic, probabilistic framework that estimates end-effector dynamics mismatches online, without relying on external force-torque sensors. From the robot's proprioceptive feedback, a Neural Network Model Mismatch Estimator generates a prior for a Variational Inference solver, which rapidly converges to the unknown parameters while quantifying uncertainty. With a real 7-DoF manipulator driven by a state-of-the-art passive impedance controller, RME adapts to sudden changes in mass and center of mass at the end-effector in $\sim400$ ms, in static and dynamic settings. We demonstrate RME in a collaborative scenario where a human attaches an unknown basket to the robot's end-effector and dynamically adds/removes heavy items, showcasing fast and safe adaptation to changing dynamics during physical interaction without any external sensory system. 

\textbf{Project Website:} \url{https://mateusz-jaszczuk.github.io/rme/}
\end{abstract}
\vspace{-10pt}
\keywords{Passive Impedance Control, Learning Residual Inverse Dynamics, Model Mismatch Estimation}
\vspace{-10pt}
\section{Introduction}
\label{sec:introduction}
\vspace{-8.5pt}
Compliant control is fundamental to the development of adaptive and collaborative robots designed to physically interact with uncertain environments, objects of unknown dynamics, and humans with unpredictable intentions and capabilities. Many commercial robots advertise \textit{compliant capabilities}, typically implemented through basic impedance or admittance control laws,  rarely possessing the ability of being \textit{passive} to physical perturbations while precise in task execution -- an essential property for fluid physical interaction \cite{energy-aware-stramigioli,BILLARD2017157}. Humans naturally achieve this balance by leveraging our redundant degrees-of-freedom \cite{redundancy-humans} and mechanical compliance in the neuro-musculoskeletal system \cite{Gottlieb1996-ia}. Replicating this behavior in robots requires control strategies that enforce a passive relationship between external forces and motion (i.e., velocity), guaranteeing stability in both free motion and during physical contact \cite{Colgate1988}. To this end, passivity-based impedance controllers have been extensively developed, guided by energy exchange techniques  \cite{Hogan1984,Stramigioli1999,Li1999,Kishi2003,Duindam2004,AlbuSchaffer2007,Ferraguti2013,kronander2015passive,Ferraguti2015,Energy2017,7801022,AbuDakka2020}; demonstrating success in a range of applications including grasping, manipulation, and robotic surgery. 
A major caveat of these techniques is that convergence and passivity guarantees depend heavily on accurate dynamics models of the robot and the environment it is interacting with. This assumption often breaks down in practice, especially when interacting with \textit{unknown environments}, such as manipulating objects with uncertain mass and inertia. In such cases, gain tuning or adding a PID layer is commonly used to mitigate model mismatch. However, these ad-hoc strategies compromise the passivity of the closed-loop system, resulting in a loss of compliance and safety.
Humans, in contrast, exhibit a remarkable ability to interact with objects of unknown dynamics and uncertain environments. For example, when lifting a box with unknown contents, we rely on an initial internal estimate of its inertial properties, and upon sensing the actual load through proprioceptive feedback, we rapidly adjust muscle activation and posture to maintain stable manipulation within our physical capabilities~\cite{Riemann2002-aq,oh_speed_2021}. We perform this rapid adjustment even when we are pushed or mistakenly bump into a wall while carrying the box. This form of rapid adaptation is key to safe and efficient physical interaction. Hence, in this work, we seek to enable robots with the ability to estimate and adapt to dynamic model mismatches, in real-time, which would allow them to maintain passivity while continuing to perform their tasks -- just like humans do. In particular, we envision scenarios in which a robot lifts a box of unknown mass, whether empty or filled with heavy books or dynamically being changed by a human, and instantaneously estimates the resulting model mismatch, enabling stable and safe manipulation while remaining passive to external physical contacts.



\textbf{Related Work} The core problem we address in this work is model mismatch, an issue that plagues model-based control. Prior works have addressed this problem, often in the context of exploration and reinforcement learning (RL) frameworks that aim to capture system dynamics accurately. For example, \citet{online_learning_legged} proposed an online learning strategy to estimate the residual dynamics model in legged locomotion based on predictive errors. The episodic learning approach proposed by \citet{self_correcting_QP} allows for robust learning of Inverse Dynamics (ID), accounting for multiple mismatch factors (internal friction, actuators' nonlinearities, model imperfections, etc.). Similarly, \citet{5509858} used Gaussian Process to improve learning Inverse Dynamics. \citet{model_plant_mismatch_compensation} proposed to compensate for model mismatch via a RL policy, learning a compensatory signal by comparing the robot's predicted and observed state transitions. Likewise, \citet{predictive_error_correction} proposed improving control policies via quantification of predictive error. Alternatively, \citet{Srour2025} proposed trajectory-optimization for manipulation under uncertain payloads. While the aforementioned approaches demonstrate robust performance, they are often sensitive to feedback measurement noise, requiring longer estimation times and thus compromising real-time performance.
Moreover, predictive error approaches may misattribute errors caused by infeasible control commands, e.g., due to unmodeled actuator limits or linearization errors, and often rely on conservative assumptions such as smooth dynamic changes starting from zero mismatch \cite{online_learning_legged}.  


\textbf{Contribution} We introduce Rapid Mismatch Estimation (RME), a probabilistic, controller and motion-planner agnostic framework for real-time quantification of dynamic model mismatch during interaction with unknown objects. RME leverages a neural network (NN) architecture trained in simulation to estimate an initial model mismatch, which is then refined with online variational inference to rapidly infer mismatches in mass and center-of-mass (CoM) at the manipulator's end-effector in $\sim$400 ms. Thus, RME provides online compensation for unknown dynamics and allows the robot to follow a desired motion policy while maintaining the passivity of the system -- just like humans do. To the best of our knowledge, no existing method provides a real-time estimation framework that can adapt to abrupt changes in model dynamics arising from interaction with unknown environments. The most related works to ours are by \citet{provably_safe_sys_id}, presenting a system-identification framework, allowing inertial parameters estimation of an unknown payload, however, requiring a 10s-long system identification/calibration action before trajectory execution, and by~\citet{com_estimation}, which introduced an effective exploration strategy for center-of-mass (CoM) estimation; however, it does not operate in real-time or during continuous task execution and depends on force/torque (FT) sensing, which we avoid as it decreases robot's dexterity, reduces allowable payload, and highly increases assembly cost~\cite{6631072}.
\vspace{-7.5pt}
\section{Problem Formulation}
\label{sec:problem_formulation}
\vspace{-7.5pt}
Let the following equation define the rigid-body dynamics of an $n$-DoF manipulator arm~\cite{spong2020},
\begin{equation} \label{eq:dynamics}
    M(q)\ddot{q}+C(q,\dot{q})\dot{q}+G(q) = \tau_c + \tau_{\text{ext}}
\end{equation}
\noindent where $\ddot{q},\dot{q},q \in\mathbb{R}^n$, represent joint accelerations, velocities, and positions, respectively. $M(q)\in\mathbb{R}^{n \times n}$, $C(q,\dot{q})\in\mathbb{R}^{n \times n}$, and  $G(q)\in\mathbb{R}^{n}$ represent the mass/inertia matrix, Coriolis matrix, and Gravity vector, respectively.  $\tau_c \in \mathbb{R}^n$ and $\tau_{\text{ext}} = J(q)^\top F_{\text{ext}}\in \mathbb{R}^n$ represent control and external torques at each joint, respectively, with $J(q)\in\mathbb{R}^{6\times n}$ being the manipulator Jacobian. 
\begin{ass}
   For robot dynamics \eqref{eq:dynamics} we can assume that $M(q)$ and $C(q,\dot{q})$ are uniformly bounded for all $q$ and $C(q,\dot{q})$ linear in $\dot{q}$. Further, we assume $C(q,\dot{q})$ fulfills the following:
   \begin{equation}
   \label{eq:skew-symmetry}
    \dot{M}(q) = C(q,\dot{q}) + C(q,\dot{q})^T  \Rightarrow   \dot{M}(q) - 2C(q,\dot{q}) ~~~ \text{skew-symmetric}
   \end{equation}
    which yields the well-known skew-symmetry condition for revolute joint robots \cite{spong2020}. 
\end{ass}
\vspace{-2.5pt}
\begin{ass}
The control torque $\tau_c$ 
in \eqref{eq:dynamics} is an impedance controller that renders a passive closed-loop behavior wrt. the environment; i.e., a passive mapping of  $(\tau_{\text{ext}} \rightarrow \dot{q})$ or $(F_{\text{ext}} \rightarrow \dot{x})$ via,
\begin{equation}
\label{eq:passivity}
    \dot{S}(q,\dot{q}) \leq \dot{q}^\top\tau_{\text{ext}} ~~~ \text{or} ~~~ \dot{S}(x,\dot{x}) \leq \dot{x}^\top F_{\text{ext}}
\end{equation}
with $S$ being the energy storage function including the kinetic energy of the robot \eqref{eq:dynamics} and the elastic potential energy from the controller. $\dot{x},F_{\text{ext}}\in\mathbb{R}^6$ is the task-space velocity and external wrench.
\end{ass}
\vspace{-5pt}
\textbf{Problem} Any passive impedance controller $\tau_c$ is generally designed so that \eqref{eq:skew-symmetry} appears in $\dot{S}$ such that \eqref{eq:passivity} can be guaranteed. This holds for controllers with feedforward/ pre-compensation terms, feedback linearization, inverse dynamics, with or without inertia shaping $M(q)$ \cite{DIETRICH2021104875,7560657,AbuDakka2020,kronander2015passive,cbf_qp_main_paper}. 
However, any interaction attempt with a heavy unknown object will result in a dynamics model mismatch. More specifically, \eqref{eq:dynamics} now includes an \textcolor{BrickRed}{additional external torque} due to model mismatch,
\begin{equation}\label{eq:dynamics_mismatch}
\underbrace{M(q)\ddot{q}+C(q,\dot{q})\dot{q}+G(q)}_{\tau_{\text{nom}}:~\text{Nominal Dynamics}}  = \tau_c + \tau_{\text{ext}} + \textcolor{BrickRed}{\underbrace{J(q)^{\top}\begin{bmatrix}
         F_m \\ r_{\text{CoM} \times F_m}
     \end{bmatrix}}_{\tau_{\text{mm}}:~ \text{Model Mismatch}}}
\end{equation}
As shown, we assume that $\textcolor{BrickRed}{\tau_{\text{mm}}}$ is caused by changing the mass properties of the end-effector (EE), modeled as a point mass added to its current inertia tensor with $F_{\text{m}}=[0,0,mg]$ being the gravitational force with $g=-9.81m/s^2$ due to unknown mass $m$ and $r_{\text{CoM}}= [r_x, r_y, r_z]$ being the unknown center-of-mass (CoM) relative to the EE. This is a valid assumption as manipulators mostly interact with the environment via the EE. Any other contact can be considered a collision or human-robot interaction, not a dynamics mismatch. Thus, if $\textcolor{BrickRed}{\tau_{\text{mm}}}$ is not compensated in $\tau_c$ this may lead to instabilities, poor tracking performance and cause unsafe behaviors for the robot and the human.

\textbf{Goal} Thus, we aim to estimate the mismatch parameters $\theta = \{m,r_x, r_y, r_z\}$ online, and explicitly compensate for any model mismatch in real-time by augmenting a passive impedance controller as,
\begin{equation}
\label{eq:rme-goal}
    \hat{\tau}_c = \tau_{c} - J(q)^{\top}\begin{bmatrix}
         \hat{F}_m(\theta) \\ \hat{r}_{\text{CoM}}(\theta) \times \hat{F}_m(\theta)
     \end{bmatrix} ~~~~ \text{with} ~~~~ \norm{ \begin{bmatrix}\hat{F}_m(\theta) - \textcolor{BrickRed}{F_m}\\
      \hat{r}_{\text{CoM}}(\theta) -  \textcolor{BrickRed}{r_{\text{CoM}}}\end{bmatrix}}^2_2 \rightarrow 0
\end{equation}
with $\tau_{c}$ being a passive impedance controller designed with the nominal dynamics \eqref{eq:dynamics}. Our choice of controller is described in Section \ref{sec:preliminaries}. As the true mismatch parameters \textcolor{BrickRed}{$F_m, r_{\text{CoM}}$ are unknown}, we seek a robust probabilistic estimate of $\theta$ by formulating an online Bayesian inference problem, as described in Section~\ref{sec:RME}. Further, we aim to estimate $\theta$ using only the robot's proprioceptive feedback (e.g., applied joint torque estimates), without a need for an external FT sensor at the end-effector. 


\vspace{-7.5pt}
\section{Preliminaries: Constrained Passive Interaction Controller \cite{kronander2015passive,cbf_qp_main_paper}}
\label{sec:preliminaries}
\vspace{-7.5pt}
Our RME framework is designed to be control-agnostic, and can be used with any passive impedance controller or even on RL trained control policies with passivity properties \cite{10610082,Zanella2024338}. In this work, however, we choose a recent extension of the dynamical systems (DS) based passive interaction controller \cite{kronander2015passive} that can guarantee passivity without any inertia shaping, training or FT sensing. In this control paradigm, the desired motion of the robot's end-effector is driven by a DS~\cite{dsbook},
\begin{equation}
\label{eq:ds}
    \dot{x}=f(x) \quad \text{s.t.} \quad \dot{\mathcal{V}}(x) = \nabla \mathcal{V}(x)^\top f(x) \leq 0
\end{equation}
which represents a motion policy that is trained (or defined) to be globally asymptotically stable (g.a.s.) wrt. an attractor $x^*\in\mathbb{R}^3\times SO(3)$ or a task-space trajectory $\{x_t\}_{t=1}^T$ like a limit cycle. The asymptotic stability of $f(x):\mathbb{R}^3\times SO(3)\rightarrow\mathbb{R}^6$ is enforced from a corresponding Lyapunov function $\mathcal{V}(x):\mathbb{R}^3\times SO(3)\rightarrow\mathbb{R}$ which can also be learned. For a comprehensive review of imitation learning (IL) of DS motion policies, see \cite{dsbook, hu_fusion_2024}. To track \eqref{eq:ds} while guaranteeing closed-loop passivity, we use the following velocity-based passive impedance control law~\cite{kronander2015passive},
\begin{equation} \label{eq:passive_impedance}
    F_{c}(x)=G_x(q)-D(x)\big(\dot{x}-f(x)\big) ~~~ \text{with} ~~~~ D(x) \succ 0
\end{equation}
with task-space gravity vector $G_x(q)\in\mathbb{R}^6$, damping matrix $D(x)\in\mathbb{R}^{6\times 6}$, and desired velocity from $f(x)$. By aligning $D(x)$ to the direction of $f(x)$ we can generate kinetic energy in the desired direction of motion while dissipating energy in the tangential direction. While this ensures passivity wrt. $F_{\text{ext}}$ via \eqref{eq:passivity}, it does not ensure that a feasible $\tau_{c} = J(q)^\top F_{c}$ satisfies any kinematic constraints, such as joint limits, self-collision, external collisions, and singularities. To address this, \citet{cbf_qp_main_paper} proposed to constrain \eqref{eq:passive_impedance} with a Quadratic Programming (QP) based safety filter, that optimizes $\tau_c$ to track $F_{c}(x)$, \begin{wrapfigure}{r}{.465\textwidth}
\vspace{-18pt}
    \begin{align}
    \label{eq:cpic}
        & \min_{\tau_c} \quad \norm{J(q)^{-\top}\tau_c - F_{c}(x) }_2^2 \quad\\
        \text{s.t} ~~~ 
        & M(q)\ddot{q}+C(q,\dot{q})\dot{q}+G(q) = \tau_c + \tau_{\text{ext}} \nonumber \\ 
        & \ddot{h}_i(q) \ge -\mathcal{K}_i[h_i(q),\dot{h}_i(q)]^\top ~~ \forall i=1,\dots \nonumber
    \end{align}
\vspace{-25pt}
\end{wrapfigure} subject to the robot dynamics \eqref{eq:dynamics} and joint kinematic constraints formulated as exponential control barrier function (CBF) with relative degree-2 \cite{nguyen2016exponential}. $h_i(q)\geq 0$ denote the invariant sets corresponding to different kinematic constraints and $\mathcal{K}_i = [k_1\ k_2]$ feedback gain vectors that render forward invariance \cite{cbf_qp_main_paper}. $J(q)^{-\top}$ is the pseudo-inverse of $J(q)^\top$. We use \eqref{eq:cpic} as the nominal passive impedance controller generating $\tau_c$ in \eqref{eq:rme-goal}. \begin{rem}
    Both the unconstrained \eqref{eq:passive_impedance} and constrained \eqref{eq:cpic} passive interaction controllers are proven to track $f(x)$ and be passive wrt. input-output pair $(F_{ext},\dot{x})$ using the energy storage function $\mathcal{S} = \frac{1}{2}\dot x^\top M_x \dot x + \lambda_1 \mathcal{V}(x)$ where $M_x$ is the task-space inertia matrix,  $\lambda_1$ is the first eigenvalue of $D(x)$ from \eqref{eq:passive_impedance} and $\mathcal{V}(x)$ the Lyapunov function used to shape $f(x)$. These guarantees rely on the satisfaction of condition \eqref{eq:skew-symmetry}. Thus, neglecting model mismatches not only degrades the tracking performance of $f(x)$ but also compromises the passivity of the closed-loop system.
\end{rem}
\begin{figure}[!tbp]
\centering
\includegraphics[width=1\linewidth]{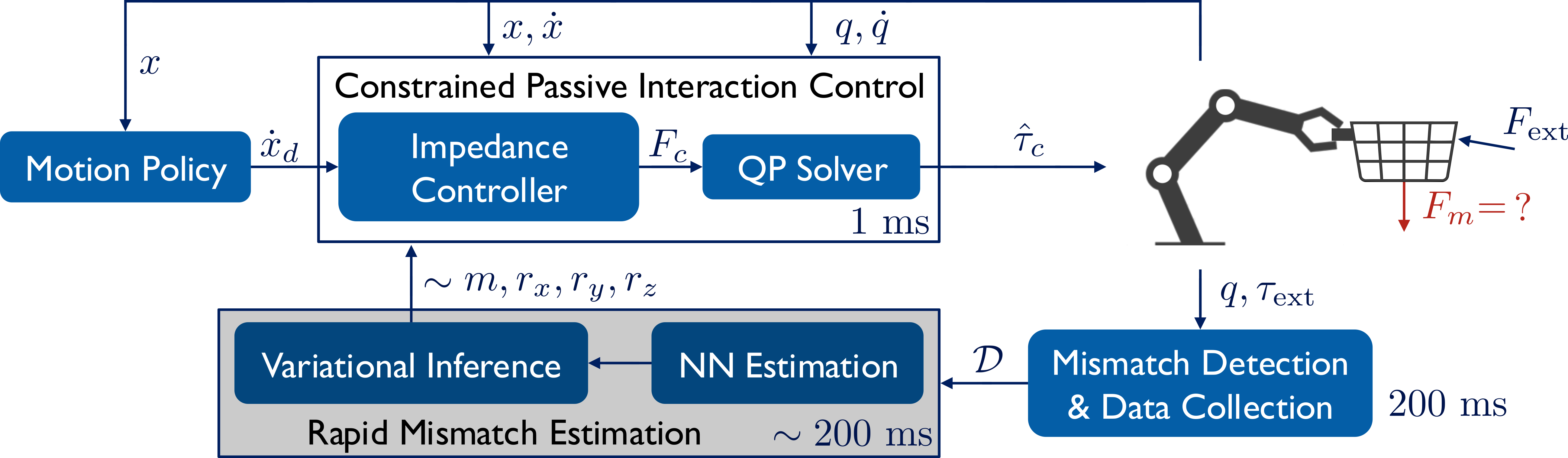}
\caption{\label{fig:flowchart} \textbf{Rapid Mismatch Estimation (RME) Framework}. The gray block denotes our novel contribution. Note that RME is controller agnostic and can be implemented with any impedance controller. In this work, we choose the constrained passive interaction controller \cite{cbf_qp_main_paper}.}
\vspace{-10pt}
\end{figure}
\vspace{-10pt}
\section{Rapid Mismatch Estimation}
\label{sec:RME}
\vspace{-7.5pt}
In order to robustly estimate the mismatch parameters $\theta = \{m,r_x, r_y, r_z\}$ while minimizing mismatch errors $\Delta F_m = \textcolor{BrickRed}{F_m} - \hat{F}_m \rightarrow \mathbf{0}$ and $\Delta r_{\text{CoM}} = \norm{\textcolor{BrickRed}{r_{\text{CoM}}} - \hat{r}_{\text{CoM}}}^2_2\rightarrow 0$ we turn to Bayesian inference. 
Rather than estimating a discrete point estimate of $\theta$, we propose to estimate a posterior distribution of mismatch parameters $p(\theta\mid\mathcal{D})$ given external torque measurements, $\tau_{\text{ext}}$, at particular joint configurations $q$ while quantifying prediction uncertainty as follows,
\begin{equation} \label{eq:eq_bayes}
    \begin{aligned}
    p(\theta \mid \mathcal{D})&=\frac{p(\mathcal{D}\mid\theta)p(\theta)}{p(\mathcal{D})},   & 
    \mathcal{D} &= \left \{ 
    \tau_{\text{ext},i} \in \mathbb{R}^n \mid i=1, \dots,N \right \}. 
    \end{aligned}
\end{equation}
\textbf{Mismatch Parameter Estimation as a Bayesian Inference Problem} Given a dataset $\mathcal{D}$ of $\tau_{\text{ext}}$ measurements collected online over a short time-window of length $N$, we seek to estimate the posterior distribution $p(\theta\mid\mathcal{D})$. We define the likelihood function with the Gaussian Distribution as the probability of observing $\tau_{\text{ext}}$ on each joint given the mismatched parameters $\theta$, computed using the manipulator arm's Inverse Dynamics (ID) model $f_{\text{ID}}(\cdot)$ with the likelihood noise $\sigma^2_{\text{likelihood}}$ based on the amplitude of noise observed at each joint in the dataset $\mathcal{D}$. 
\begin{equation}
\label{eq:p_functions}
    \begin{aligned}
    p(\mathcal{D}|\theta)& = \mathcal{N}(f_{\text{ID}}\left(q,\theta),\operatorname{diag}(\sigma^2_{\text{likelihood}})\right),   & 
    p(\theta)&=\mathcal{N}\left(f_{\text{NN}}(\mathcal{D}),\operatorname{diag}(\sigma^2_{\text{prior}})\right)
    \end{aligned}
\end{equation}
This inference problem does not have a unique solution in the short time window, as multiple mismatch parameters will result in similar $\tau_{\text{ext}}$ observations, making CoM estimation particularly sensitive to the choice of the prior $p(\theta)$. Therefore, we propose to model $p(\theta)$ using the prediction of a NN $f_{\text{NN}}(\cdot)$ as the mean for the Gaussian Distribution, as described in Section~\ref{sec:nn}. Notice that, even though we have \eqref{eq:p_functions} the true posterior distribution $p(\theta \mid \mathcal{D})$ is still intractable, as the manipulator's dynamics exhibit non-linear dependencies on mismatch parameters $\theta$. We address this by employing a Variational Inference (VI) framework, under which we aim to construct a mean-field approximation to the true posterior~\cite{VI_review, VI_auto_encoding}, guided by the NN $f_{\text{NN}}(\cdot)$ initial solutions.

\textbf{RME Framework} As shown in Figure~\ref{fig:flowchart}, our framework begins with mismatch detection algorithm that constantly monitors $\norm{\tau_{\text{ext}}}^2_2$, distinguishing dynamics mismatch from human-generated perturbations by checking for rapid increases followed by short-term stabilization at a high value, characteristic for rapidly added loads, allowing the RME to run when only the dynamics mismatch is detected. Implementation details of our mismatch detector are provided in Appendix~\ref{app:activation_condition}. Once a mismatch is detected, we record the dataset $\mathcal{D}$ of $\tau_{\text{ext}}$ measurements over a 200 ms time window while the robot is being controlled by $\tau_c$. $\mathcal{D}$ is then fed into the initial NN mismatch estimator, $f_{\text{NN}}(\cdot)$, which will predict the mean value for the prior as defined in \eqref{eq:p_functions}. We then run the VI framework to approximate the posterior $p(\theta\mid\mathcal{D})$ and rapidly converge to the parameter estimate $\theta$. 
\begin{figure}[!tbp]
\centering
\includegraphics[width=0.95\linewidth]{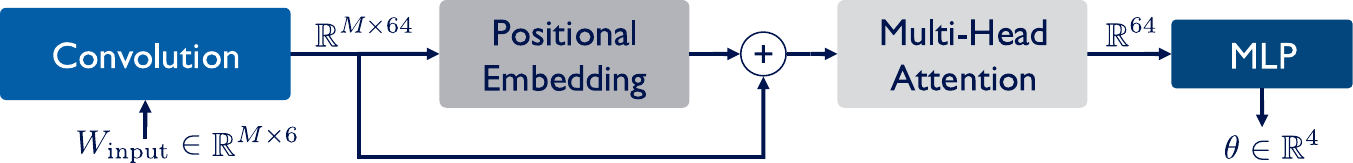}
\caption{\label{fig:nn_architecture} \textbf{RME Neural Network Architecture}. In the network, we input a sequence of pseudo-wrenches of dimension $\mathbb{R}^{M\times 64}$, apply a convolution layer, positional embedding, and multi-head attention. Further, we mean pool over the attention scores and apply a sequential Multilayer Perceptron, which performs a final regression to mismatch parameters $\theta$.}
\vspace{-15pt}
\end{figure}
\vspace{-5pt}
\subsection{Neural Network Model Mismatch Estimator Architecture}\label{sec:nn}
\vspace{-5pt}
In this work, we chose to construct and train a NN to guide VI by using it as a meaningful mean prior $p(\theta)$. The NN architecture, depicted in Figure~\ref{fig:nn_architecture}, was designed to analyze the $M$-dimensional sequence of tokens, representing time-series of external pseudo-wrenches, $\hat{W}_{\text{ext}}\in\mathbb{R}^6$, applied to the EE, computed using the external torque measurements at given joint configurations, with damped pseudo-inverse of the Jacobian matrix, where $\lambda$ is a damping coefficient, as below. 
\begin{equation}
    \hat{W}_{\text{ext}}= \left( J(q)J(q)^{\top} + \lambda I \right)J(q) \tau_{\text{ext}}, \ W_{\text{input}} = \left \{ 
    \hat{W}_{\text{ext,i}} \in \mathbb{R}^6 \mid i=1, \dots,M \right \}
\end{equation}
The sequence of inputs to the NN is then $W_{\text{Input}}\in\mathbb{R}^{M\times 6}$, where $M$ inputs are uniformly sampled from $N$ collected data points. To solve the regression problem of the mismatch parameters, maintaining translation-invariance relative to the input sequence, 
\begin{equation}
    \left \{q_{i} \in \mathbb{R}^n, \ \tau_{\text{ext},i} \in \mathbb{R}^n \mid i=1, \dots,M \right \} \longrightarrow W_{\text{Inputs}} \xrightarrow{\text{NN}} \{m,r_x, r_y, r_z\}
\end{equation}
we design an architecture that keeps the forward pass computationally efficient. First, we apply a 1D convolution~\cite{LeCun_cnn} with kernel size 5 and output dimension 64, to capture local patterns in the data, enhancing translation-invariance. Then, we apply position embedding and multi-head attention~\cite{attention} with 8 attention heads, to capture global dependencies between all tokens and situate them in the sequence. Further, we apply the mean polling of attention outputs over the sequence length to obtain a vector of dimension $\mathbb{R}^{64}$ that we pass to a sequence of three MLP blocks, each consisting of a linear layer $(64 \rightarrow256)$, ReLU activation function, and a linear layer $(256 \rightarrow 64)$. Then, we apply a final linear layer $(64 \rightarrow4)$ to estimate the mean mismatch parameters $\theta$ for $p(\theta)$. 

We train this NN mismatch estimator in simulation, on a dataset generated by 350 simulations of a manipulator's dynamics experiencing different mismatch parameters. During training, we use Mean-Squared Error loss objective, and apply dropout, with dropout rate $d=0.1$ after each of the MLP blocks, to prevent network overfitting~\cite{dropout}. We also employ data augmentation to train the NN to be robust to noise and can transfer sim-to-real. Details on the architecture and training are in Appendix~\ref{app:nn_architecture}. We provide ablations results on NN impact on the inference in Section~\ref{sec:evaluation}. 
\vspace{-5pt}
\subsection{Variational Inference for Mismatch Estimation Convergence}\label{sec:vi}
\vspace{-5pt}
During runtime, the $f_{\text{NN}}(\cdot)$ trained in the previous sections is used to construct $p(\theta)$ as in \eqref{eq:p_functions} to start the VI framework. Since the true posterior distribution $p(\theta \mid \mathcal{D})$ is intractable, we construct a mean-field approximation to the true posterior $p(\theta \mid \mathcal{D})\approx q_{\phi}(\theta \mid \mathcal{D})$, parameterized by the following variational parameters $\phi$~\cite{VI_review, VI_auto_encoding},
\begin{equation}
    \phi = (\underbrace{\mu_m, \ \mu_x, \ \mu_y, \ \mu_z,}_{\mu_{\phi}} \  \underbrace{\sigma^2_m, \ \sigma^2_x, \ \sigma^2_y, \ \sigma^2_z}_{\sigma^2_{\phi}} )
\end{equation}
We formulate an optimization objective for computing the variational parameters $\phi$ that yield an effective approximation of the true posterior by minimizing the Kullback-Leibler divergence.
\begin{equation}\label{eq:eq_kl}
    q_{\phi}^*(\theta \mid \mathcal{D}) = \operatorname{arg} \min_{q_{\phi}} \mathrm{KL} \left[ q_{\phi}(\theta \mid \mathcal{D}) \parallel p(\theta \mid \mathcal{D}) \right]
\end{equation}
Under the mean-field approximation, \eqref{eq:eq_kl} can lead to a tractable optimization objective, by the maximization of the Evidence Lower Bound (ELBO), which is equivalent to the minimization of KL-divergence, that we optimize via stochastic gradient descent~\cite{VI_auto_encoding,ADVI},
\begin{equation}
    \text{ELBO}(\phi) = \mathcal{H}[q_{\phi}(\theta|\mathcal{D})]+\mathbb{E}_{\theta \sim q_{\phi}(\theta|\mathcal{D})}[\log p(\mathcal{D},\theta)]
\end{equation}
with $\mathcal{H}[q_{\phi}(\theta|\mathcal{D})]$ being the entropy of $q_{\phi}(\cdot)$.
\begin{equation}
    \mathcal{L}(\phi) = -\text{ELBO}(\phi) = -\mathbb{E}_{\theta \sim q_{\phi}(\theta|\mathcal{D})}[\log p(\mathcal{D}|\theta) + \log p(\theta) - \log q_{\phi}(\theta | \mathcal{D})]
\end{equation}
\begin{equation}
\label{eq:vi-obj}
    \phi^* =  \operatorname{arg} \min_{\phi} \ \mathcal{L}(\phi)
\end{equation}
The expectation of the ELBO loss can be effectively computed with Monte Carlo estimation via the reparameterization trick~\cite{VI_auto_encoding},
\begin{equation}
    \begin{aligned}
    \theta_i &\overset{iid}{\sim} q_{\phi}(\theta \mid \mathcal{D}), & 
    \theta_i &= \mu_\phi + \sigma_{\phi} \odot \varepsilon, &
    \varepsilon &\sim \mathcal{N}(0,I_4)
    \end{aligned}
\end{equation} 
which allows us to evaluate gradients of \eqref{eq:vi-obj} with automatic differentiation~\cite{ADVI}. To estimate $\phi$, we use the Adam optimizer~\cite{Adam} with learning rate $\eta=0.025$ and gradient clipping to prevent gradient explosion in early optimization stages. We run the optimizer until variational parameters stabilize at a given threshold, allowing us to return the optimization result as soon as it converges without performing unnecessary training iterations, accelerating the estimation process. Note that inaccurate NN prediction will not impact convergence, as it only affects the prior $p(\theta)$.
\vspace{-5pt}
\subsection{Preservation of Passivity and Stability of the Closed-Loop System}
\label{sec:preservation_passivity_stability}
\vspace{-5pt}
One of the main motivating factors for developing RME was to achieve a fast enough estimation of $\theta$ such that we could preserve the task convergence and passivity of the closed-loop system provided by the passive interactive controllers introduced in Section~\ref{sec:preliminaries}. Note that, when our estimated mismatch parameters are near perfect, $\Delta F_m \approx \mathbf{0}$ and $\Delta r_{\text{CoM}} \approx 0$, and we control the robot dynamics \eqref{eq:dynamics_mismatch} with our proposed augmented controller $\hat{\tau}_c$ as defined in \eqref{eq:rme-goal} then the mismatch and compensation terms cancel out and we recover the original dynamics \eqref{eq:dynamics} whose closed-loop behavior is stable and passive under a control input $\tau_c$ defined by \eqref{eq:passive_impedance} or \eqref{eq:cpic} as proven in \cite{kronander2015passive, dsbook} and \cite{cbf_qp_main_paper}, respectively.

\begin{pro}
    Let a robotic manipulator with dynamics \eqref{eq:dynamics_mismatch} be controlled by an augmented passive impedance control law defined in \eqref{eq:rme-goal} with $\tau_c$ computed by \eqref{eq:cpic}. Given an imperfect mismatch parameter estimation $\Delta F_m \neq \mathbf{0}$ and $\Delta r_{\text{CoM}} \neq 0$ the system will be locally asymptotically stable as unwanted equilibria may arise when $\norm{\Delta F_\text{mm} - F_c(x)} = 0$, with $\Delta F_\text{mm}$ denoting the remaining unknown model mismatch in the EE expressed as a task-space wrench. Yet, the closed-loop system behavior remains passive wrt. input-output port $(F_{\text{ext}} + \Delta F_{\text{mm}},\dot{x})$.\\
    \textbf{Proof:} Provided in Appendix~\ref{app:proof}.\hfill $\blacksquare$
\end{pro}

\section{Evaluation}
\label{sec:evaluation}
\vspace{-5pt}
We evaluate our model with physical robot experiments using a 7-DoF Franka Emika torque-controlled manipulator arm with embedded torque sensors. We empirically tuned the standard deviation for the VI prior as $\sigma_{\text{prior}} = \begin{bmatrix}0.5& 0.02& 0.02& 0.05\end{bmatrix}$. The QP optimization for \eqref{eq:cpic} was implemented using CVXGEN~\cite{CVXGEN}; the controller and RME were evaluated on a workstation with $11^{\text{th}}$ Gen Intel\textsuperscript{\textregistered} Core\textsuperscript{TM} i7-11700K @ 3.60 GHz CPU. No GPU acceleration is required.

\vspace{-10pt}
\subsection{Ablations on NN Effect and Hyper-Parameters for VI}
\label{sec:effect_of_nn}
\vspace{-5pt}
We analyze the effect of NN on VI prediction accuracy by comparing inference results achieved with $p(\theta)$ constructed using $\mu_{\text{prior}}=f_{\text{NN}}(\cdot)$ and $\mu_{\text{prior}}=0$. We ran the RME algorithm on 100 simulation-generated datasets, representing the manipulator's dynamic behavior under different mismatch parameters $\theta$. As shown in Table~\ref {tab:ablation_mse}, NN guidance significantly reduces the Mean Squared Error in CoM estimation. Detailed predictions comparison is shown in Appendix~\ref{app:ablation_study_nn}. We further analyze the impact of the estimation interval length on inference accuracy in Appendix~\ref{app:ablation_interval_length}, showing that 200 ms is an optimal length for fast and robust estimation, making the solver more robust to measurement noise while allowing not smooth but abrupt changes, causing noticeable and instantaneous disruption in the trajectory. Since we can solve the estimation problem in $\sim$ 400 ms, we can run this framework sequentially, ensuring that if the estimation was biased by the external factors, like undetected perturbation, it can be corrected by the subsequent prediction.
\begin{table}[!htbp]
\renewcommand{\arraystretch}{1.25}
\centering
\begin{tabular}{|c|c|c|c|c|}
 \hline
 $\mu_{\text{prior}}$ & MSE $m$ $(kg^2)$ & MSE $r_x$ $(m^2)$ & MSE $r_y$ $(m^2)$ & MSE $r_z$ $(m^2)$ \\
\hline
$0$ & 0.319 $\times 10^{-3}$ & 0.594 $\times 10^{-3}$& 0.880$\times 10^{-3}$ & 6.513$\times 10^{-3}$\\
\hline
$f_{\text{NN}}(\cdot)$ & 0.229 $\times 10^{-3}$& 0.366 $\times 10^{-3}$& 0.629$\times 10^{-3}$ & 2.560$\times 10^{-3}$\\
\hline
\end{tabular}
\vspace{0.25cm}
\caption{Mean Squared Errors for RME estimation of mismatch parameters $\theta$ with Gaussian Prior $p(\theta)$ constructed by $\mu_{\text{prior}}=f_{\text{NN}}(\cdot)$ and $\mu_{\text{prior}}=0$.}
\label{tab:ablation_mse}
\vspace{-20pt}
\end{table}
\subsection{Estimation Results from Static Experiments}
\label{sec:est_results}
\vspace{-5pt}
To evaluate RME, we extensively tested the model with multiple static experiments, where the physical manipulator, subject to sudden changes in the dynamics model resulting from adding unknown mass to the end-effector, aimed to maintain the target equilibrium position and orientation using the constrained passive impedance controller (CPIC)~\eqref{eq:cpic}. RME provided rapid and accurate estimation of mismatch parameters $\theta$, as shown in Figure~\ref{fig:parity}, and further in Appendix~\ref{app:detailed_estimation_results}, with an average estimation time of 226 ms. As shown in Figure~\ref{fig:static_tracking_error}, adaptation allowed the robot to quickly converge to the equilibrium position from before applying the mismatch - note that even if the tracking error does not reduce to zero, due to disturbance or under-prediction, RME always provides valid compensation, resulting in trajectory correction.
\begin{figure}[!htbp]
\vspace{-5pt}
  \begin{minipage}[b]{.475\linewidth}
    \centering
    \includegraphics[width=0.85\linewidth]{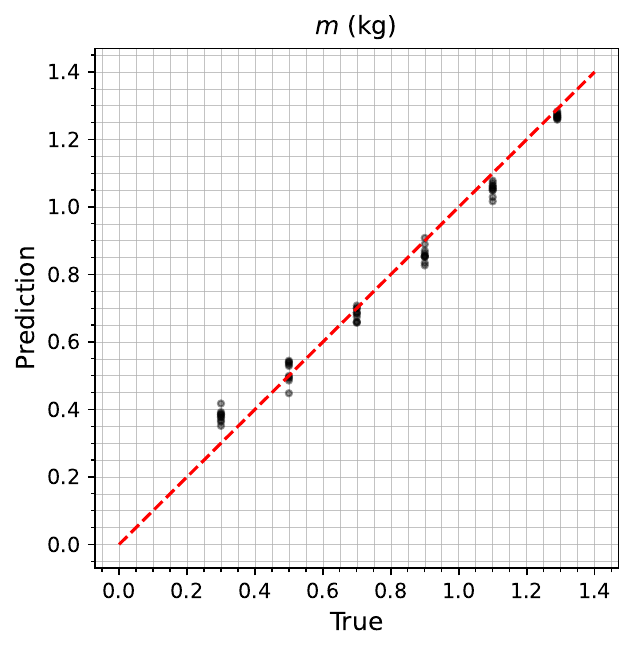}
    \caption{\label{fig:parity}Parity plot for mass predictions over 60 independent experiments. Results show consistent and stable mass estimation using RME.}
  \end{minipage}\hfill
  \begin{minipage}[b]{.475\linewidth}
\centering
    \includegraphics[width=0.9\linewidth]{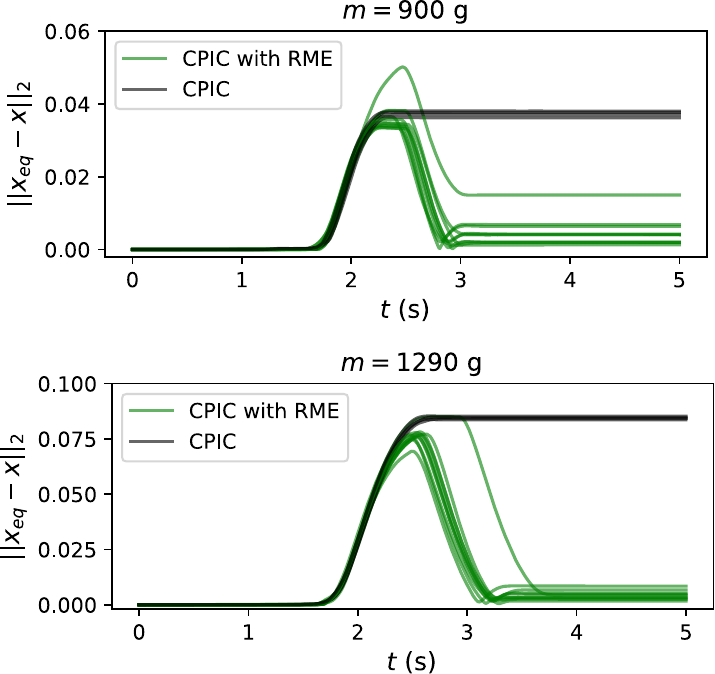}
    \caption{\label{fig:static_tracking_error}Franka manipulator arm tracking performance subject to mismatch in the dynamics model applied at $t\approx1.75$ s.}
  \end{minipage}
  \vspace{-15pt}
\end{figure}

Under this setup, we observed that when the CoM of applied mismatch is close to the z-axis of the end-effector, as in experiments with 1.1 kg and 1.29 kg, the CoM offset causes negligible twist of the end-effector, reducing its observability due to non-linear dependencies of Inverse Dynamics on mismatch parameters $\theta$, shown in results in Appendix~\ref{app:detailed_estimation_results}. However, the robot still safely converges to the target state, and CoM can be corrected in subsequent predictions if the twist increases due to a change in orientation.
\newpage
\subsection{Performance on Dynamic Experiments}
\label{sec:dynamic_experiments}
\vspace{-5pt}
\begin{wrapfigure}{r}{.345\textwidth}
\vspace{-15pt}
    \includegraphics[width=0.9\linewidth]{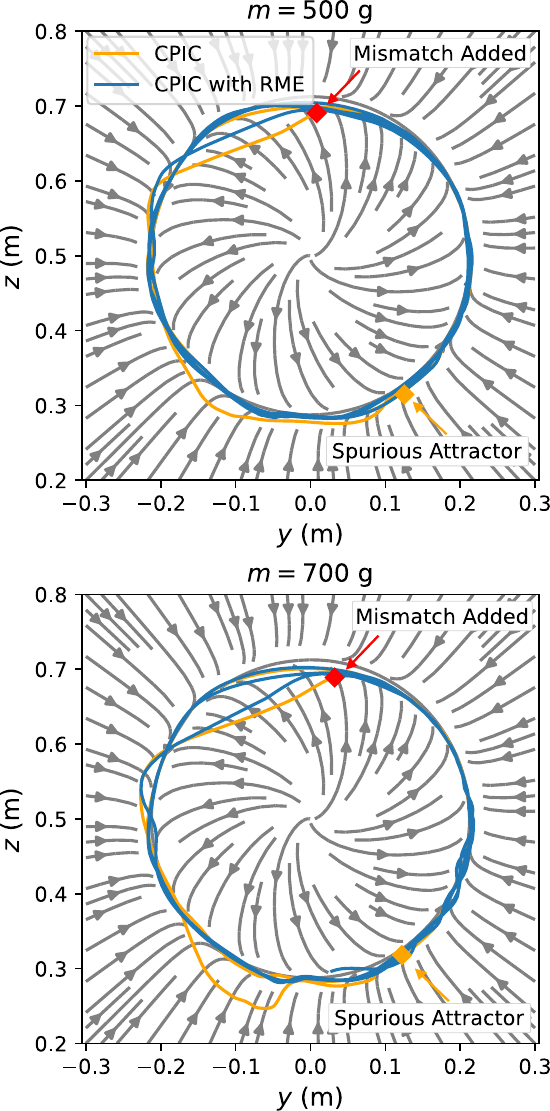}
\caption{\label{fig:limit_cycle}Manipulator tracking a DS with a stable limit cycle in the y-z plane subject to mismatch in the dynamics model.}
\vspace{-25pt}
\end{wrapfigure}\textbf{Tracking a Stable Limit Cycle}
To test the model in the dynamic scenario, we designed a Dynamical System with a stable limit cycle in the y-z plane, where the manipulator aims to achieve target velocities from the motion planner. As shown in Figure~\ref{fig:limit_cycle}, RME rapidly estimates mismatch parameters, allowing the robot to converge to the desired trajectory. Without adaptation, the manipulator reaches the spurious attractor, where the task force from the impedance controller is balanced by the gravitational force acting on the mass mismatch, preventing it from continuing on the given task.

\begin{figure}[!htbp]
\centering
\includegraphics[width=1.0\linewidth]{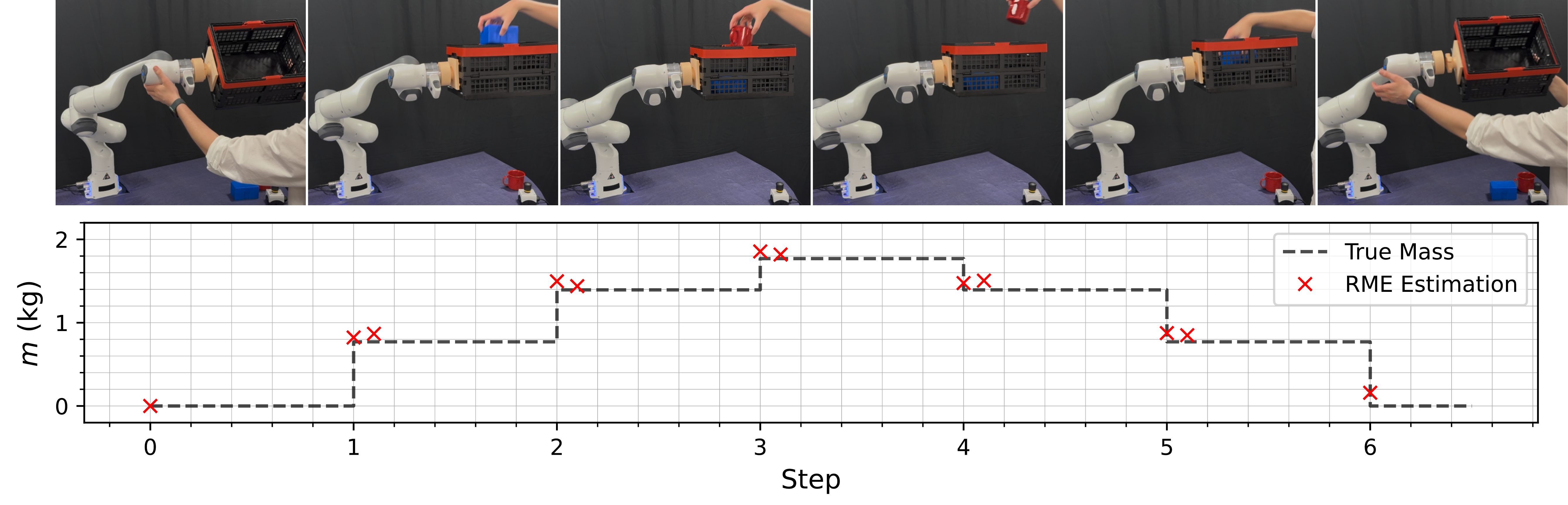}
\vspace{-18pt}
\caption{\label{fig:basket_experiment}Manipulator adaptation to sequential changes in the dynamics model, while subject to human-generated perturbations. Each step represents modifying the dynamics model of the end effector; RME prediction between two steps denotes immediate correction of the previous $\theta$ estimate.}
\vspace{-20pt}
\end{figure}

\textbf{Sequential Adaptation with Human-Robot Interactions} In this experiment, we evaluate the model's capability to perform continuous adaptation while maintaining passivity with respect to human-generated perturbations. We attach the unknown basket to the end effector, sequentially add and remove two unknown objects, and finally, remove the basket, perturbing the robot between these actions. RME provided an accurate estimation of mismatch parameters $\theta$, allowing the robot to maintain target position and orientation, while being passive to human-generated forces (even when the model confused human perturbations with model mismatch, it allowed perturbation, and instantaneously corrected itself after being released). Figure~\ref{fig:basket_experiment} shows the sequential adaptation's experimental setup and mass estimation history. Additional pHRI experiments are provided in Appendix~\ref{app:Pick_and_place}.
\section{Conclusion}
\label{sec:conclusion}
\vspace{-5pt}
We propose a novel adaptation framework capable of estimating mismatch in the dynamics model of the impedance-controlled manipulator in $\sim$ 400 ms. We evaluated our work in a series of experiments, where we abruptly changed the dynamics model of the robot, and showed that our model can provide accurate estimation and compensation, allowing the robot to safely complete the task and stay passive wrt. to human-generated perturbations. In our future work, we plan to utilize uncertainty quantification to develop a robust motion policy that will allow precise estimation of the CoM, addressing the observability issue under certain configurations. We also plan to utilize uncertainty quantification and Neural Network formulation shown in Section~\ref{sec:nn} to develop a more complex mismatch detection condition, allowing more robust mismatch detection. Further, we plan to extend the RME to model more complex dynamic interactions with unknown objects and explicitly account for unmodeled actuator dynamics, improving the framework's applicability and accuracy.

\clearpage

\section{Limitations}
To enable safe human-robot interactions, it is crucial to disambiguate between human-robot interaction and mismatch in the manipulators’ dynamic model, ensuring that the framework will not compensate for human-generated perturbations. We notice that in selected scenarios, our simple activation condition might be unable to distinguish these two factors in such a short time window. Therefore, in our future work, we plan to develop and implement more
complex mismatch detection condition, allowing more robust and safe human-robot interaction.

During the evaluation, we observed that when the CoM of applied mismatch is close to the z-axis of the end-effector, its observability is reduced due to non-linear dependencies of Inverse Dynamics on mismatch parameters $\theta$. Although the robot could converge to the desired trajectory in all experiments, and CoM prediction can be corrected in the subsequent prediction, in our future work, we plan to develop an exploration strategy that utilizes prediction uncertainty to allow efficient and accurate CoM estimation. In addition, we observed that when the end-effector is experiencing rapid accelerations along the global z-axis, such as in the stable limit cycle experiment, the mismatch prediction might be biased. Further, unmodeled actuator dynamics, such as joint friction, can be reflected in the $\tau_{\text{ext}}$ measurements, also biasing the RME predictions. Although in both cases, predictions can be corrected in subsequent estimations, we plan to address these limitations in our future work, further improving the framework's accuracy.

Finally, our model assumes that the detected object is a point mass added to the current inertia tensor of the end-effector. This assumption holds when the manipulator is not rapidly accelerating, which is valid for Human-Robot interaction scenarios; however, inertia estimation can be incorporated into the framework to achieve a more accurate system representation. 
\acknowledgments{We thank our anonymous reviewers, who provided thorough and fair feedback that improved the quality of our paper. This work was supported by the National Science Foundation (NSF) Foundational Research in Robotics (FRR) program under NSF CAREER Award Grant No. FRR-2443721.}
\bibliography{references}  
\newpage

\appendix
\section*{Appendix}
\section{Mismatch Detection Algorithm}
\label{app:activation_condition}
\begin{figure}[!htbp]
\centering
\includegraphics[width=0.725\linewidth]{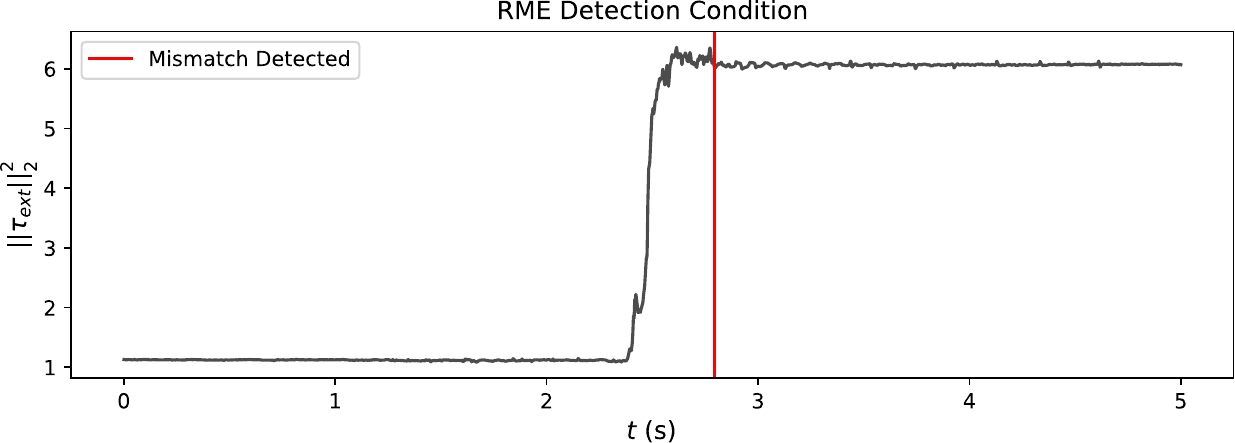}
\caption{\label{fig:detection_condition}Example of mismatch detection algorithm execution. The algorithm monitors $\norm{\tau_{\text{ext}}}^2_2$ and detects a mismatch in dynamics after observing a $\norm{\tau_{\text{ext}}}^2_2$ increase followed by stabilization within a given threshold.}
\end{figure}

\vspace{-5pt}
To ensure that our framework runs only when mismatch is detected, we designed a simple mismatch detection algorithm that constantly monitors $\norm{\tau_{\text{ext}}}^2_2$, checking for rapid increases or decrease exceeding activation threshold, which, depending on the experiment, we set between 0.7 and 1.1 Nm, followed by stabilization within 0.2 Nm threshold over 230 ms period, characteristic for applying mismatch to the end effector. Algorithm~\ref{alg:activation_codition} shows the implementation of the mismatch detection condition. The algorithm analyzes $\norm{\tau_{\text{ext}}}^2_2$ time-history over 500 ms of the most recent feedback, monitoring for rapid change of $\norm{\tau_{\text{ext}}}^2_2$. Further, it evaluates stabilization condition by computing the mean $\norm{\tau_{\text{ext}}}^2_2$ over short time windows within the stabilization interval and compares it with the mean $\norm{\tau_{\text{ext}}}^2_2$ at the end of the interval, allowing detection of both adding and removing objects from the end effector. To improve the algorithm's robustness in pHRI, we also analyze measured pseudo-wrenches applied to the end-effector, verifying that the principal applied force acts in the global z-direction, and that forces in x and y directions do not exceed 2.5 N, ensuring that mismatch is caused only by the applied load, not human perturbation. Detection threshold, stabilization threshold, and stabilization interval were found empirically through $\norm{\tau_{\text{ext}}}^2_2$ profiles analysis and can be tuned based on the task objective. This simple condition allowed us to detect a mismatch applied to the end-effector and run RME only when all detection conditions were met simultaneously. Further, the algorithm can detect when dynamics mismatch and human perturbations occur simultaneously, as these events exhibit different torque profiles, allowing the detection algorithm to distinguish the two. Hence, RME will be activated once perturbations are no longer present, preserving the controller's passivity. Figure~\ref{fig:detection_condition} shows an example of algorithm execution. After detecting a mismatch, we proceed to data collection, which is followed by the RME framework estimation.
\begin{algorithm}
\caption{Mismatch Detection Algorithm}\label{alg:activation_codition}
\begin{algorithmic}
\Function{ActivationCondition}{\{$\norm{\tau_{\text{ext}}}^2_2, \hat{F}_{\text{ext}} \}_{i=1}^N$} \Comment{N=500 steps}
\vspace{5pt}
\State $\text{rapid change} \gets \left| \norm{\tau_{\text{ext}}}_2^2 (500)-\norm{\tau_{\text{ext}}}_2^2 (1) \right|>\text{activation threshold}$
\vspace{5pt}
\State $\begin{array}{@{}c@{}}
\text{stabilization} \\
\text{checks}
\end{array} \gets \left[\frac{1}{20}\sum\limits_{i=431}^{450} \norm{\tau_{\text{ext}}}_2^2 (i), \ \frac{1}{20}\sum\limits_{i=351}^{370} \norm{\tau_{\text{ext}}}_2^2 (i), \  \frac{1}{20}\sum\limits_{i=271}^{290} \norm{\tau_{\text{ext}}}_2^2 (i) \right]$
\vspace{5pt}
\State $\text{end stabilization} \gets \frac{1}{10}\sum\limits_{i=491}^{500} \norm{\tau_{\text{ext}}}_2^2 (i)$
\vspace{5pt}
\State $\overline{F}_{\text{ext}} \gets \left[\frac{1}{50}\sum\limits_{i=451}^{500} \hat{F}_{\text{ext, x}}, \ \frac{1}{50}\sum\limits_{i=451}^{500} \hat{F}_{\text{ext, y}}, \  \frac{1}{50}\sum\limits_{i=451}^{500} \hat{F}_{\text{ext, z}}\right]$
\vspace{5pt}
\State $\text{principal force check} \gets \left(|\overline{F}_{\text{ext, z}}|>|\overline{F}_{\text{ext, x}}| \right) \land \left(|\overline{F}_{\text{ext, z}}|>|\overline{F}_{\text{ext, y}}| \right)$
\vspace{5pt}
\State $\text{side forces check} \gets \left(|\overline{F}_{\text{ext, x}}|<2.5 \right) \land \left(|\overline{F}_{\text{ext, y}}|<2.5 \right)$
\vspace{5pt}
\State $\text{force check} \gets \text{principal force check} \  \land \ \text{side forces check}$
\vspace{5pt}
\State $\text{stabilization} \gets \textbf{True}$
\vspace{5pt}
\ForAll{$\text{checks} \in \ \text{stabilization checks}$}
        \If{$\left|\text{checks} - \text{end stabilization}\right| \geq \text{stabilization threshold}$}
            \State $\text{stabilization} \gets$ \textbf{False}
        \EndIf
    \EndFor
\vspace{5pt}
\State $\text{mismatch detection} \gets \text{rapid change} \  \land  \ \text{stabilization} \ \land \ \text{force check} $
\State \Return $\text{mismatch detection}$
\EndFunction
\end{algorithmic}
\end{algorithm}
\vspace{-20pt}

\newpage
\section{Compensatory Action Formulation}
\label{app:QP_controller}
\vspace{-5pt}
To compensate for the estimated mismatch, we define $\hat{\tau}_c$~\eqref{eq:rme-goal} as control torque $\tau_c$ from the Constrained Passive Interaction Controller with additional mismatch gravity compensation term. In our formulation, we solve the CPIC~\eqref{eq:cpic} subject to equality constraints from the Nominal Dynamics model, not corrected by predicted model mismatch. We found that this strategy mitigates potential instabilities due to RME overcompensation, while succeeding in trajectory tracking. This approach also maintains RME controller-agnostic, as the compensation is provided without influencing the nominal controller. However, any impedance controller used with RME can be corrected using the estimation result to improve tracking accuracy further. 

\section{Effect of Neural Network on Inference Accuracy}
\label{app:ablation_study_nn}
\begin{figure}[!htbp]
\centering
\includegraphics[width=0.95\linewidth]{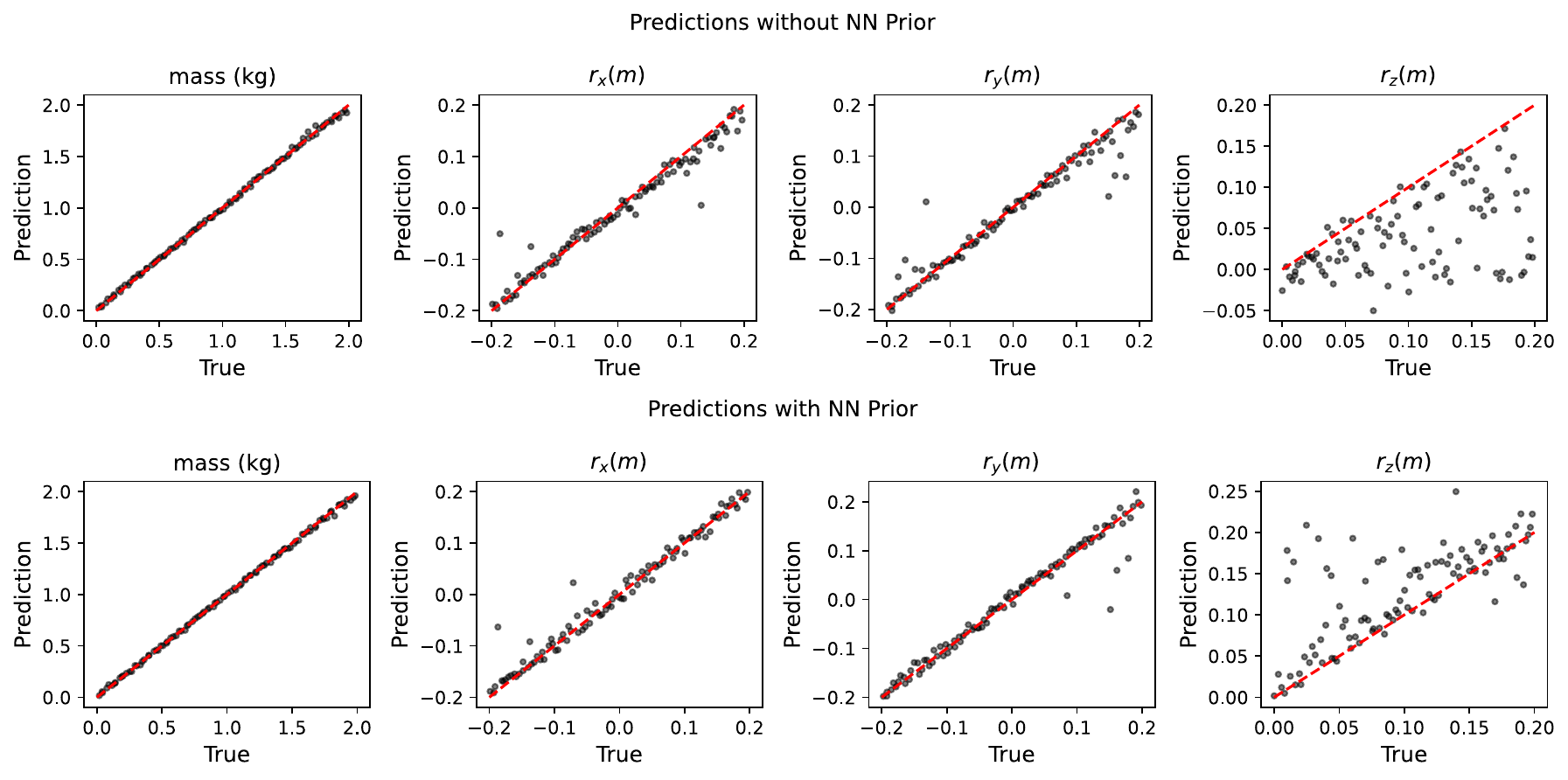}
\caption{\label{fig:ablation_study}Comparison of parity plots for RME estimation of mismatch parameters $\theta$ under different prior constructions; as shown, Neural Network guidance allows for more robust center of mass estimation.}
\end{figure}

To examine the importance of the Neural Network in our framework, we conducted an ablation study, comparing inference results achieved with $p(\theta)$ constructed by $\mu_{\text{prior}}=f_{\text{NN}}(\cdot)$ and $\mu_{\text{prior}}=0$. By running the RME algorithm on 100 simulation-generated datasets, representing the manipulator's dynamic behavior under different mismatch parameters $\theta$, we compare RME estimation results under different prior constructions, as shown in Figure~\ref{fig:ablation_study}. Since Inverse Dynamics is non-linearly dependent on mismatch parameters $\theta$, Neural Network allows us to guide the Inference toward the true center of mass estimation, allowing for more accurate mismatch parameters $\theta$ estimation.

\section{Pick and Place with Human-Robot Interactions}
\label{app:Pick_and_place}
\begin{figure}[!htbp]
\centering
\includegraphics[width=0.98\linewidth]{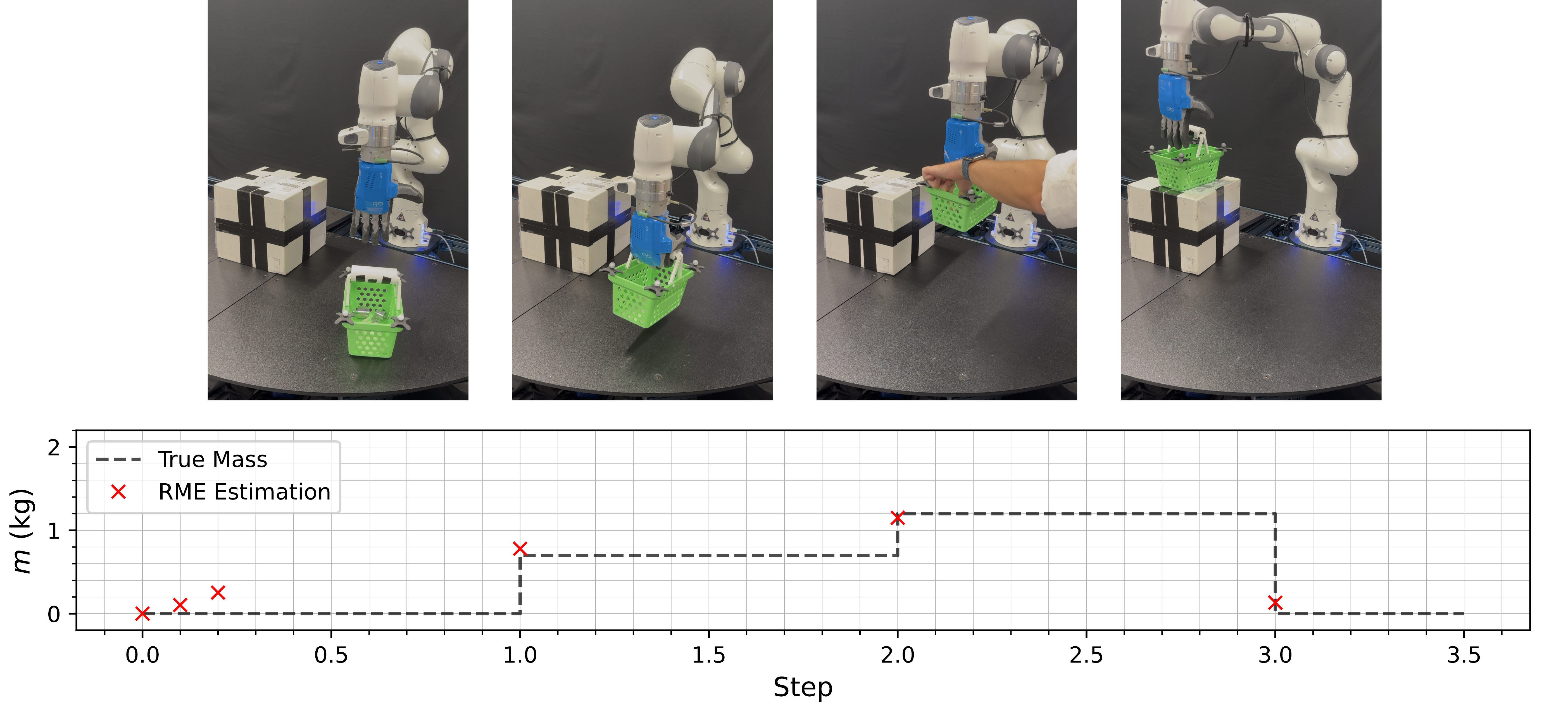}
\caption{\label{fig:pick_place}Manipulator adaptation to sequential changes in the dynamics model while interacting with unknown, heavy objects. Each step represents modifying the dynamics model of the end effector; at Step 1, the manipulator picks up an unknown 700-gram basket, at Step 2, the user adds 500 grams to the basket, and at Step 3, the robot places the basket on top of the box. RME prediction between two steps denotes immediate correction of the previous $\theta$ estimate.}
\end{figure}

\begin{figure}[!htbp]
\centering
\includegraphics[width=0.98\linewidth]{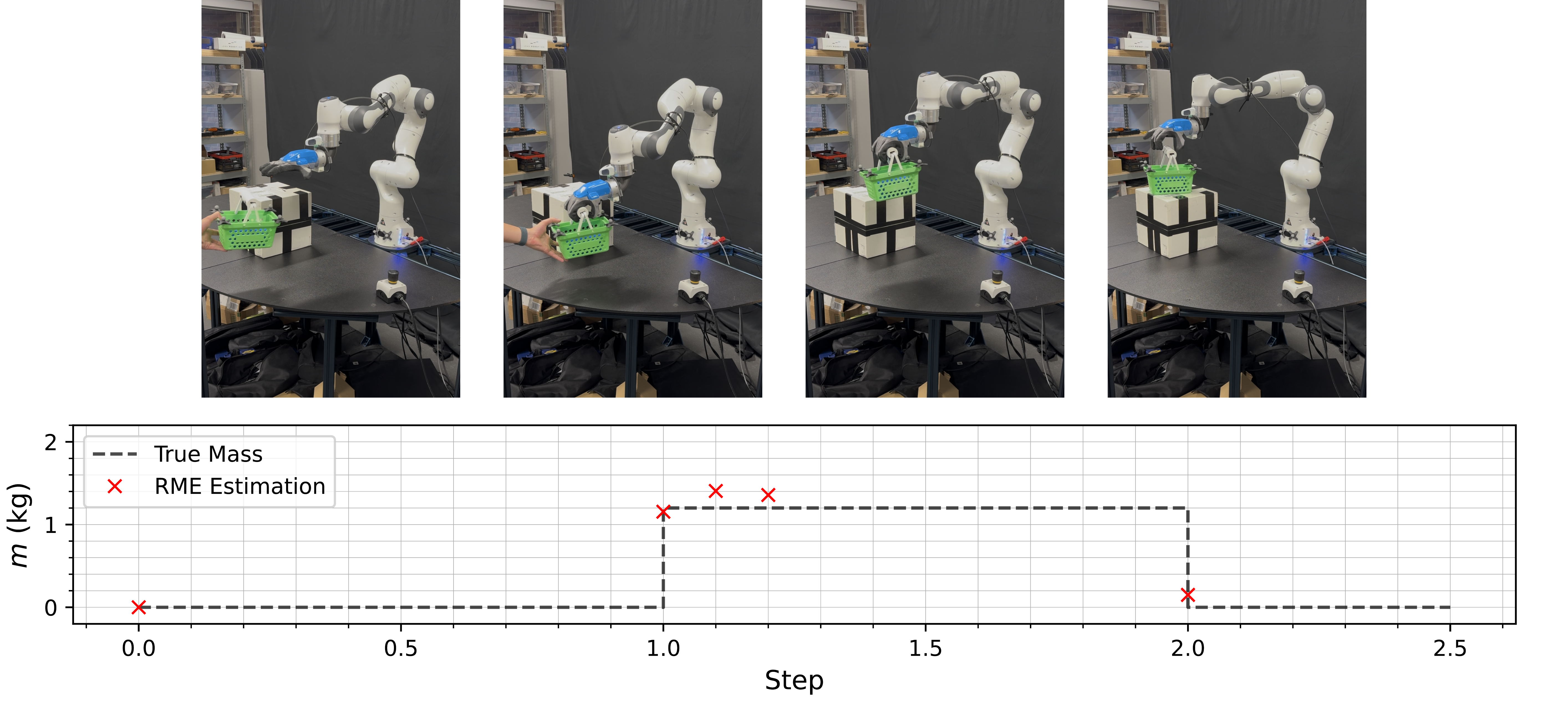}
\caption{\label{fig:intercept_basket}Manipulator intercepting basket of unknown mass (1200 grams) from a user and placing it on top of the box. Using RME, the robot can safely adapt and finish the given task, while maintaining passivity wrt. human-generated perturbations, ensuring safe pHRI. RME prediction between two steps denotes immediate correction of the previous $\theta$ estimate.}
\end{figure}

\vspace{-5pt}
In this experiment, we evaluate model performance in more complex continuous adaptation scenarios by performing a pick and place task with a passive velocity-based inverse kinematics (PVIK) controller \cite{mirrazavi2018sca-dual}, utilizing the qb SoftHand2 Research end effector and the OptiTrack motion capture system to track the basket. In the first test, Figure~\ref{fig:pick_place}, the robot aims to pick up an unknown object (700 grams) and place it on top of the box, maintaining passivity with respect to human-generated perturbations. During the process, we add 500 grams to the basket. RME allows the robot to track the desired trajectory by rapidly adapting to unknown dynamics, while a controller without RME fails to pick up a heavy object. In the second test, Figure~\ref{fig:intercept_basket}, the robot aims to track and intercept the basket (1200 grams) from a human and place it on top of the box, which is only possible with the use of RME. In both experiments, the controller with RME augmentation preserves stability and passivity of the closed-loop system.

\section{RME Estimation Results from Static Experiments}
\label{app:detailed_estimation_results}
\begin{figure}[!htbp]
\centering
\includegraphics[width=0.8\linewidth]{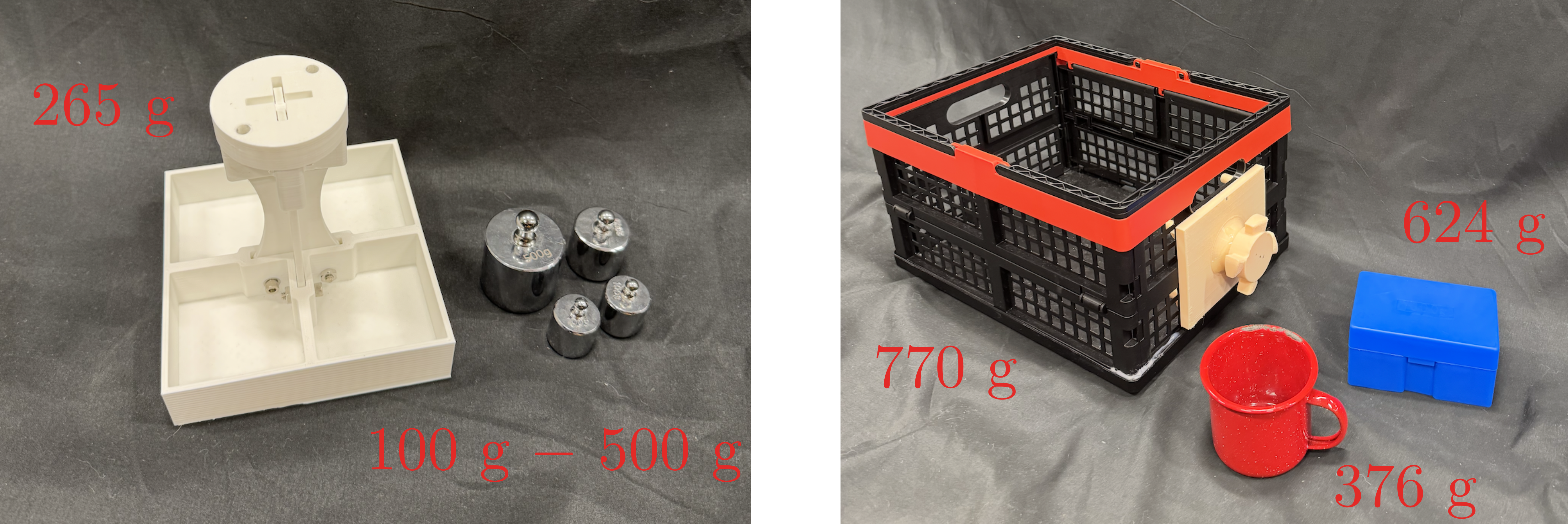}
\caption{\label{fig:experimental_end_efectors}End-effectors and unknown objects used in the RME evaluation. The left image shows the 3D-printed basket used for static experiments and limit cycle tracking. The right image depicts a commercially available basket with a 3D printed connector, used in a Sequential Adaptation with Human-Robot interactions experiment.}
\end{figure}

To perform the evaluation presented in Section~\ref{sec:evaluation}, we designed two end effectors as baskets attached to the manipulator, as shown in Figure~\ref{fig:experimental_end_efectors}. The white end-effector was 3D printed and served as the base for static experiments and limit cycle tracking. Its construction allowed us to place heavy objects at precise positions, providing reliable ground truth information for RME evaluation. Further, we designed a 3D printed connector to a commercially available basket to test Sequential Adaptation with Human-Robot Interactions. This setup allowed us to verify RME adaptation capabilities with everyday objects, ensuring our model is generalizable to various tasks. 

Table~\ref{tab:framework_predictions} shows RME estimation results over 60 independent experiments, where the physical manipulator, subject to sudden changes in the dynamics model resulting from adding unknown mass to the end-effector, aimed to maintain target equilibrium position and orientation, as described in Section~\ref{sec:est_results}. Estimation results are mean predictions and standard deviations from 10 experiments per Applied Mismatch. The average RME estimation time over 60 experiments was 226 ms.

\begin{table}[!htbp]
\renewcommand{\arraystretch}{1.25}
\setlength{\tabcolsep}{5pt}
\centering
\begin{tabular}{|c|c|c|c|c|c|c|c|}
\hline
 \multicolumn{4}{|c|}{\textbf{Applied Mismatch}} & \multicolumn{4}{c|}{\textbf{Mean Prediction}} \\
 \hline
  $m$ (kg) & $r_x$ (m) & $r_y$ (m) & $r_z$ (m) & $m$ (kg) & $r_x$ (m) & $r_y$ (m) & $r_z$ (m) \\
\hline
0.300  & 0.03 & 0.00 & 0.13 & 0.380 $\pm$0.018 & -0.04$\pm$0.03 & 0.02$\pm$0.01 & 0.13$\pm$0.04 \\
\hline
0.500  & 0.06 & 0.00 & 0.13 & 0.511$\pm$0.031 &  0.04$\pm$0.04 & 0.01$\pm$0.01 & 0.15$\pm$0.02 \\
\hline
0.700  & 0.06 & 0.00 & 0.13 & 0.681$\pm$0.019 &  0.07$\pm$0.03 & 0.00$\pm$0.01 & 0.14$\pm$0.02 \\
\hline
0.900  & 0.05 &-0.03 & 0.13 & 0.861$\pm$0.024 &  0.07$\pm$0.01 & -0.02$\pm$0.01 & 0.11$\pm$0.03 \\
\hline
1.100  & 0.05 & 0.02 & 0.13 & 1.053$\pm$0.018 & 0.07$\pm$0.01 & 0.00$\pm$0.01 & 0.00$\pm$0.04 \\
\hline
1.290  & 0.05 & 0.01 & 0.13 & 1.269$\pm$0.008 &  0.05$\pm$0.00 & 0.00$\pm$0.01 & 0.00$\pm$0.06 \\
\hline
\end{tabular}
\vspace{0.25cm}
\caption{Framework mean estimation results and standard deviations over 60 independent experiments. At each trial, we recorded the robot's proprioceptive feedback over a 200-ms-long time window, followed by the model estimation, which took 226 ms on average. In each experiment, RME prediction allowed the robot to converge back to the desired trajectory.}
\label{tab:framework_predictions}
\end{table}

\begin{figure}[!htbp]
\centering
\includegraphics[width=0.9\linewidth]{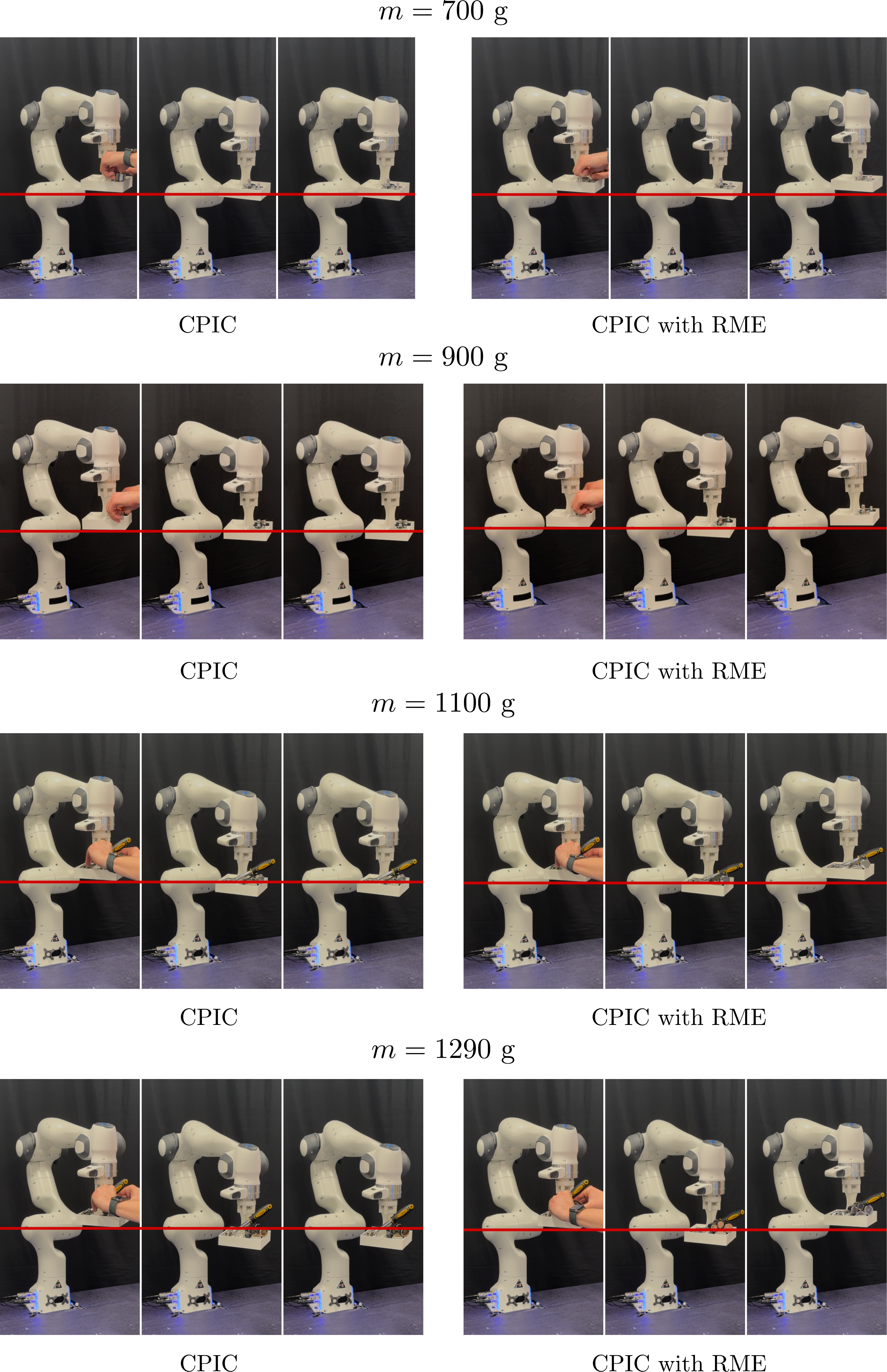}
\vspace{0.15cm}
\caption{\label{fig:static_experiments_motion}Comparison of manipulators' behavior subject to unknown mismatch in dynamics model ranging from 700 g to 1290 g, when controlled with Constrained Passive Interaction Controller (CPIC), and CPIC with RME adaptation. Red lines indicate the manipulator's equilibrium position before experiencing the mismatch. As shown, for all weights used in the experiment, RME allows the robot to correct its position and orientation, resulting in convergence to the equilibrium position; CPIC-controlled manipulators deviate from the equilibrium state and fail to converge to the goal.}
\end{figure}

As mentioned in Section~\ref{sec:est_results}, the manipulator's goal is to maintain target equilibrium position and orientation when interacting wth unknown objects from the environment. Examples of such adaptations are depicted in Figure~\ref{fig:static_experiments_motion}. As shown, the robot controlled with CPIC without the estimation framework deviates from the target position and orientation as it is pulled by the additional mass attached to the end-effector. RME allows the manipulator to correct its position and converge to the equilibrium point from before applying the mismatch.

\section{Effect of Data Collection Interval Length on Inference Accuracy}
\label{app:ablation_interval_length}
 To choose the data collection interval for the Neural Network and Variational Inference, we conducted a comparative analysis to examine model performance under different collection intervals. We evaluated the study with 11 datasets from real robot experiments to ensure the model can account for measurement noise and external factors not present in the simulation environment. As shown in Table~\ref {tab:ablation_data_interval}, the RME model achieves lower MSE values with increased collection intervals. Although we noticed that a longer collection interval might allow us to capture more information-rich feedback, based on this study, we decided to set the collection interval to 200 ms, as it allows us to achieve high estimation accuracy while shortening RME execution time. The collection interval length can be tuned based on the particular manipulator and task specifications. 
\begin{table}[!htbp]
\renewcommand{\arraystretch}{1.25}
\centering
\begin{tabular}{|c|c|c|c|c|}
 \hline
Collection Interval & MSE $m$ $(kg^2)$ & MSE $r_x$ $(m^2)$ & MSE $r_y$ $(m^2)$ & MSE $r_z$ $(m^2)$ \\
\hline
50 ms & 2.480 $\times 10^{-3}$ &1.607 $\times 10^{-3}$ &0.574 $\times 10^{-3}$ &8.935$\times 10^{-3}$\\
\hline
100 ms &2.436$\times 10^{-3}$ & 0.909 $\times 10^{-3}$ & 0.468$\times 10^{-3}$ & 4.827$\times 10^{-3}$\\
\hline
200 ms &2.467$\times 10^{-3}$ & 0.86 $\times 10^{-3}$ & 0.396$\times 10^{-3}$ & 3.88$\times 10^{-3}$\\
\hline
300 ms &2.081 $\times 10^{-3}$ &0.979 $\times 10^{-3}$ &  0.477 $\times 10^{-3}$ &1.25$\times 10^{-3}$\\
\hline
\end{tabular}
\vspace{0.25cm}
\caption{Comparison of RME Mean Squared Errors for $\theta$ parameters for estimation with different data collection intervals.}
\label{tab:ablation_data_interval}
\end{table}

\section{Neural Network Architecture}
\label{app:nn_architecture}
The summary of the Neural network architecture described in Section~\ref{sec:nn} is shown in Table~\ref{tab:NN_architecture}. As input, we choose a sequence of $M=20$ pseudo-wrenches $\hat{W}_{\text{ext}}$ calculated using the manipulator's proprioceptive feedback, uniformly sampled from $N$ collected data points over a 200 ms time window. Decreasing input dimensionality $(N=200 \rightarrow M=20)$ reduces model complexity without negatively impacting model performance. 
\begin{figure}[!htbp]
\centering
\includegraphics[width=0.85\linewidth]{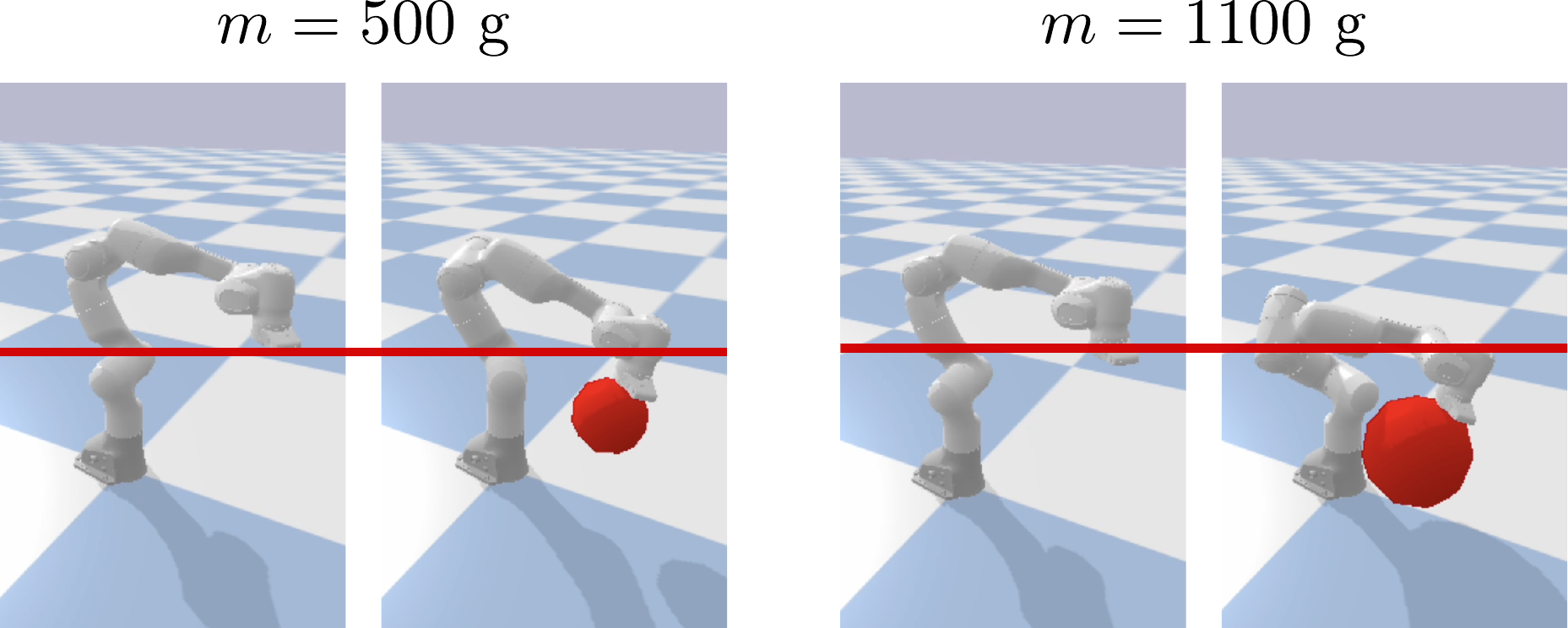}
\caption{\label{fig:sim_mismatch}Computer simulations of the manipulator's dynamics subject to an unknown mass attached to the end effector. The figure compares the manipulator's equilibrium position before applying the mismatch (red line) to its position shortly after applying unknown masses of 500 grams and 1100 grams (red spheres), shown in the left and right figures, respectively.}
\end{figure}

\begin{figure}[!htbp]
\centering
\includegraphics[width=1.0\linewidth]{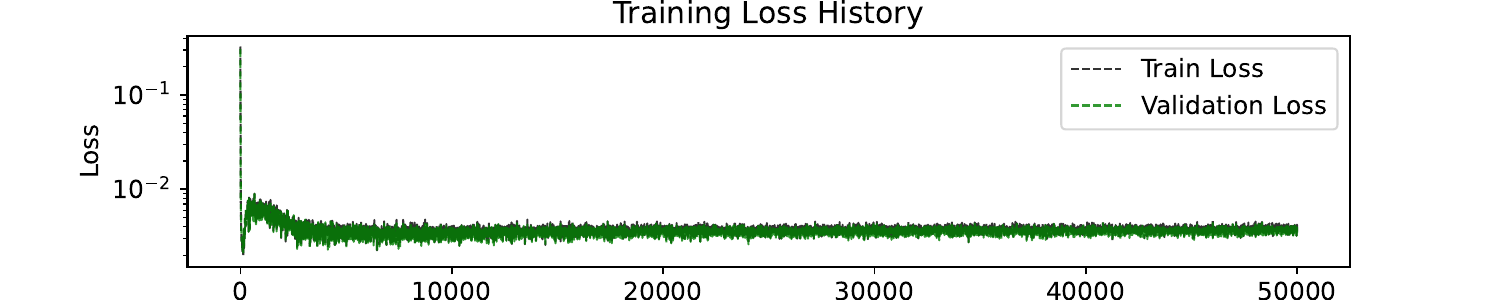}
\caption{\label{fig:loss}Neural Network train and validation loss history.}
\end{figure}

To train the Neural Network, we used a dataset generated from 350 computer simulations of a manipulator's dynamics experiencing different mismatch parameters $\theta$, as shown in Figure~\ref{fig:sim_mismatch}. This allowed us to train the NN without performing multiple experiments on the physical robot, making the framework easily generalizable to other manipulator arms. We further augmented the dataset by extracting input time series beginning at different time steps. This provided us with 1050 training inputs, which we split into 80 - 20 training-validation datasets. We tested the network performance with a separate dataset representing time-translated data relative to the training dataset; the NN made accurate predictions of mismatch parameters, which is essential, assuming we cannot guarantee the exact positioning of the extracted time series of data. Figure~\ref{fig:loss} shows the training and validation loss history over the training iterations. As shown in the plot, both loss curves converge, suggesting that the applied dropout prevented the model from overfitting. 

\begin{table}[!htbp]
\renewcommand{\arraystretch}{1.25}
\centering
\begin{tabular}{|c|c|}
 \hline
 Parameter & Value\\
\hline
Input Sequence Length $(M)$ & 20 \\
\hline
Input Dimension & $M\times 6$ \\
\hline
Convolution Kernel Size & 5 \\
\hline
Convolution Output Dimension & $M\times 64$ \\
\hline
Number of Attention Heads & 8 \\
\hline
MLP Hidden Dimension & 256 \\
\hline
MLP Activation Function & ReLU \\
\hline
MLP Sequential Blocks & 3 \\
\hline
Dropout Rate & 0.1 \\
\hline
Output Dimension & 4\\
\hline
Total Number of Parameters & 53,252 \\
\hline
Train-Validation Split & 80 -- 20\\
\hline
Training Dataset Size & $M\times 6 \times 840$\\
\hline
Validation Dataset Size & $M\times 6 \times 210$\\
\hline
Learning Rate & 0.0001\\ 
\hline
Training Iterations & 50,000 \\
\hline 
Training Time & 1h 0m 26s\\
\hline
\end{tabular}
\vspace{0.25cm}
\caption{Summary of the Neural Network architecture.}
\label{tab:NN_architecture}
\end{table}

\newpage
\section{Proof of Proposition 4.1}
\label{app:proof}
Let us re-state the proposition that must be proven in this section:

\textbf{Proposition 4.1} Let a robotic manipulator with dynamics \eqref{eq:dynamics_mismatch} be controlled by an augmented passive impedance control law defined in \eqref{eq:rme-goal} with $\tau_c$ computed by \eqref{eq:cpic}. Given an imperfect mismatch parameter estimation $\Delta F_m \neq 0$ and $\Delta r_{\text{CoM}} \neq 0$ the system will be locally asymptotically stable as unwanted equilibria may arise when $\norm{\Delta F_\text{mm} - F_c(x)} = 0$, with $\Delta F_\text{mm}$ denoting the remaining unknown model mismatch in the EE expressed as a task-space wrench. Yet, the closed-loop system behavior remains passive wrt. input-output port $(F_{\text{ext}} + \Delta F_{\text{mm}},\dot{x})$. \hfill $~~\blacksquare$ 
\begin{proof} Without loss of generality, assuming a single constraint exists, we can convert the CPIC optimization problem \eqref{eq:cpic} enacting the E-CBF kinematic constraint to \cite{cbf_qp_main_paper}:
\begin{equation}
\label{eq:cpic_single}
\begin{aligned}
        & \min_{\tau_c} \quad \norm{J(q)^{-\top}\tau_c - F_{c}(x) }_2^2 
        & \text{s.t} ~~~\begin{cases}
        M(q)\ddot{q}+C(q,\dot{q})\dot{q}+G(q) = \tau_c + \tau_{\text{ext}}  \\ 
        \nabla_qh(q)^\top\ddot{q} \ge b(q,\dot{q})
        \end{cases}
\end{aligned}
\end{equation}
given $\mathcal{K} = [k_1 ~ k_2]$ and $b(q,\dot{q}) = -k_1h(q) - k_2\nabla_qh(q)^{\top}\dot q - {\dot q}^{\top} \nabla^2_q h(q) \dot{q}$. In \cite{cbf_qp_main_paper}, the authors mention that if an optimal $\tau_c^*$ for \eqref{eq:cpic_single} can be found in the feasible set, i.e., when $J(q)^{-\top}\tau_c =F_c(x)$ or $F_c \notin \mathcal{N}(J(q)^{-\top})$  then the closed-loop behavior of the robot is \textit{passive when feasible}. Next, we prove this statement in a more rigorous fashion. Let us rewrite the equality constraint in \eqref{eq:cpic_single} as:
\begin{equation}
\label{eq:dynamics_acc}
\begin{aligned}
\ddot{q} = M(q)^{-1} (\tau_c + \rho(q, \dot{q})) ~~ & \text{with} ~~ \rho(q, \dot{q}) := \tau_{\text{ext}} - C(q, \dot{q})\dot{q} - G(q)
\end{aligned}
\end{equation}
Via \eqref{eq:dynamics_acc} the CPIC controller \eqref{eq:cpic_single} can be expressed as:
\begin{equation}
\begin{aligned}
        & \min_{\tau_c} \quad \norm{J(q)^{-\top}\tau_c - F_{c}(x) }_2^2 
        & \text{s.t.} ~~~ 
        \nabla_qh(q)^\top\left(M(q)^{-1} (\tau_c + \rho(q, \dot{q})) \right) \ge b(q)
\end{aligned}
\end{equation}
which can be re-written in terms of $\tau_c$ as follows:
\begin{equation}
\label{eq:qp_standard}
\begin{aligned}
\min_{\tau_c} \quad \frac{1}{2} \tau_c^\top Q \tau_c + \tau_c^\top p  + c \quad \text{s.t.} \quad A \tau_c \geq \nu
\end{aligned}
\end{equation}
where $c=||F_c(x)||_2^2$ and we define the following variables: 
\begin{equation}
    \begin{aligned}
     Q := 2J(q)^{-1} J(q)^{-\top} \quad & \quad  A := \nabla_q h(q)^\top M(q)^{-1}\\
     p := -2 J(q)^{-1} F_c(x) \quad &  \quad  \nu := b(q) - A\rho(q, \dot{q})
    \end{aligned}
\end{equation}
This shows that the CPIC controller \eqref{eq:cpic_single} is reduced to a standard constrained QP, which is always feasible. Hence, we can express the optimal solution $\tau_c^*$ as an analytic closed-form expression via KKT conditions. The KKT system for the \textit{active inequality} constraints becomes:
\begin{equation}
\label{eq:kkt_active}
    \begin{bmatrix}
        Q & A^\top\\
        A & 0 
    \end{bmatrix}    \begin{bmatrix}
        \tau_c^*\\
        \lambda^* 
    \end{bmatrix} = \begin{bmatrix}
        -p\\
        \nu 
    \end{bmatrix}
\end{equation}
with $\lambda^*$ being the optimal solution of  Lagrange multipliers corresponding to the inequality constraints $A\tau_c\geq\nu$ in the Lagrangian of \eqref{eq:qp_standard} which is $\mathcal{L}(\tau_c,\lambda) = \frac{1}{2} \tau_c^\top Q \tau_c + \tau_c^\top p  - \lambda^\top(A \tau_c - \nu)$. 

We will next derive the closed-form expression in the following two scenarios. 

\textbf{When the robot is far from violating the kinematic constraints,} this means that the E-CBF inequality $\nabla_qh(q)^\top\ddot{q} - b(q,\dot{q}) \ge 0 $ \underline{is satisfied}, in such cases the inequality constraint is \textit{inactive} and thus the optimal solution to \eqref{eq:qp_standard} (via \eqref{eq:kkt_active}) becomes:
\begin{equation}
\label{eq:tauc_unconstr}
\begin{aligned}
    \tau_c^* & = -Q^{-1}p\\
    & = \left(2 J(q)^{-1} J(q)^{-\top}\right)^{-1}2 J(q)^{-1} F_c(x)\\
    & = J(q)^{\top} J(q) J(q)^{-1} F_c(x) \quad \quad \quad \text{(assuming $J(q)$ is full rank)}\\
    & = J(q)^{\top}F_c(x).
\end{aligned}
\end{equation}
which reduces to the solution of the original passive interaction controller 

\textbf{When the robot is close to violating (or in violation) of the kinematic constraints,} this means that the E-CBF inequality  $\nabla_qh(q)^\top\ddot{q} - b(q,\dot{q}) \geq 0 $ is \underline{not satisfied}, in such cases an optimal QP solution, $\tau_c^*$ and $\lambda^*$ can be expressed by the analytical solution of the KKT system \eqref{eq:kkt_active} as below:
\begin{equation}
\label{eq:tauc_constr}
\begin{aligned}
    \tau_c^* & = Q^{-1} \left(-p -A^\top \lambda^*\right) \\
    & = J(q)^{\top}F_c(x) - Q^{-1} A^\top \lambda^*
\end{aligned}
\end{equation}
assuming $J(q)$ is full rank and with: 
\begin{equation}
    \lambda^* = \left( A Q^{-1} A^\top \right)^{-1} \left( A Q^{-1} p + \nu \right)\\
\end{equation}
shows that the constrained solution is a correction to the unconstrained torque estimate $J(q)^{\top}F_c(x)$ from \eqref{eq:tauc_unconstr}. During execution, the solution of the CPIC \eqref{eq:cpic_single} will either be \eqref{eq:tauc_unconstr} or \eqref{eq:tauc_constr}.

Thus, we can now analyze the closed-loop stability and passivity of these two solutions when plugged into the robot dynamics \eqref{eq:dynamics_mismatch} via our augmented controller with RME compensation \eqref{eq:rme-goal} as, 
\begin{equation}
\label{eq:dynamics-rme-error}
	M(q)\ddot{q} + C(q, \dot{q})\dot{q} + G(q) = \tau_{c}^* - J(q)^{\top}(F_{ext} + \Delta F_{mm}),
\end{equation}
with $\Delta F_{mm}$ representing the estimation error of the \textcolor{BrickRed}{true unknown model mismatch}:
\begin{equation}
    \Delta F_{mm} = \begin{bmatrix}
         \hat{F}_m(\theta) \\ \hat{r}_{\text{CoM}}(\theta) \times \hat{F}_m(\theta)\end{bmatrix} - \textcolor{BrickRed}{\begin{bmatrix}
         F_m \\ r_{\text{CoM} \times F_m}
     \end{bmatrix}}
\end{equation}
As our control and RME compensation is defined in task-space we convert \eqref{eq:dynamics-rme-error} to, 
\begin{equation}\label{eq:dynamics-error-cart}
	M_x(q)\ddot{x} + C_x(q, \dot{q})\dot{x} + G_x(q) = J(q)^{-\top}\tau_{c}^* + (F_{ext} + \Delta F_{mm}),
\end{equation}
with the gravity vector being mapped to task-space by $G_x(q) = J(q)^{-\top} G(q)$, and $M_x(q) = J(q)^{-\top} M(q) J(q)^{-1}$,  
$C_x(q, \dot{q})= J(q)^{-\top} C(q, \dot{q}) J^{-1} - J(q)^{-\top} M(q) J(q)^{-1} \dot{J}(q) J(q)^{-1}$. Similarly, the joint-space velocities are mapped to task-space as $\dot{x}=J(q)\dot{q}$ with $\dot{x}\in\mathbb{R}^d$ and joint-space accelerations $\ddot{q}$ mapped to task-space as follows $\ddot{x} = J(q)\ddot{q} + \dot{J}(q)\dot{q}$. 

\textbf{Closed-loop behavior when $\tau_c^*$ solved as \eqref{eq:tauc_unconstr}} Let us now analyze the closed-loop behavior of \eqref{eq:dynamics-error-cart} when the robot is not close to a kinematic constraint; i.e., $\tau_c^*$ is expressed as \eqref{eq:tauc_unconstr}, which reduces to:
\begin{equation}
\begin{aligned}
    	M_x(q)\ddot{x} + C_x(q, \dot{q})\dot{x} + G_x(q) & = J(q)^{-\top}J(q)^{\top}\underbrace{F_c(x)}_{\eqref{eq:passive_impedance}} + (F_{ext} + \Delta F_{mm})\\
         M_x(q)\ddot{x} + C_x(q, \dot{q})\dot{x}  & = -D(x)\big(\dot{x}-f(x)\big) + (F_{ext} + \Delta F_{mm})\\
\end{aligned}
\end{equation}
The damping function $D(x)$ can be designed to dissipate energy in orthogonal directions to $f(x)$ formulated as $D(x) = V(x)\Lambda(x)V(x)^T$ with (for a 2D example): 
\begin{equation}
\label{D-ortho}
\begin{aligned}
V(x) = \begin{bmatrix} e_1(x) & e_2(x) \end{bmatrix}, \quad
e_1(x) = \frac{f(x)}{\|f(x)\|}, \quad e_1(x)^T e_2(x) = 0, \quad \Lambda(x) = \begin{bmatrix}
\lambda_1(x) & 0 \\
0 & \lambda_2(x)
\end{bmatrix}
\end{aligned}
\end{equation}
with the eigenvalues $\lambda_i(x)\geq 0$ setting the impedance in orthogonal directions of $Q(x)$ basis, as originally presented \cite{kronander2015passive}. These $D(x)$ design choices for the controller simplify it to:
\begin{equation}
\label{eq:closed-loop-rme-uncontrained}
         M_x(q)\ddot{x} + \left(C_x(q, \dot{q}) + D(x)\right) \dot{x}  - \lambda_1f(x)  =  (F_{ext} + \Delta F_{mm})
\end{equation}
The stability and passivity of the robot's closed-loop dynamics with RME \eqref{eq:closed-loop-rme-uncontrained} can be proven by following \eqref{eq:passivity} with the following energy storage function: 
 \begin{equation}
 \label{eq:storage-cons}
     S(x,\dot{x}) = \frac{1}{2}\dot{x}^TM_x\dot{x} + \lambda_1 \mathcal{V}(x)
 \end{equation}
that includes the kinetic energy of the robot and the potential energy injected by the controller term. The latter depends solely on the eigenvalue $\lambda_1$ and the Lyapunov function $\mathcal{V}(x)$ used to ensure asymptotic stability of $f(x)$, which for this analysis we assume to be conservative $f(x)=-\nabla\mathcal{V}(x)$. The time-derivative of the energy storage function is, 
\begin{equation}
 \label{eq:power-cons}
  \begin{aligned}
\dot{S}(x,\dot{x}) & =  \dot{x}^TM_x\ddot{x} + \frac{1}{2}\dot{x}^T\dot{M}_x\dot{x} + \lambda_1\nabla\mathcal{V}^T\dot{x}\\
& = \dot{x}^T\left((F_{ext} + \Delta F_{mm}) - (C_x + D(x))\dot{x} + \lambda_1f(x)\right) + \frac{1}{2}\dot{x}^T\dot{M}_x\dot{x} + \lambda_1\nabla\mathcal{V}(x)\\
& = \dot{x}^T(F_{ext} + \Delta F_{mm}) - \dot{x}^TD(x)\dot{x} + \frac{1}{2}\dot{x}^T\underbrace{(\dot{M}_x-2C_x)}_{=0 ~~\eqref{eq:skew-symmetry}}\dot{x} + \lambda_1 \left(f(x) + \nabla\mathcal{V}(x)\right)\\
& = \dot{x}^T(F_{ext} + \Delta F_{mm}) - \dot{x}^TD(x)\dot{x}\\
& \leq \dot{x}^T(F_{ext} + \Delta F_{mm})
\end{aligned}  
\end{equation}
Thus, the following conditions are guaranteed, 
\begin{equation}
\label{eq:guarantees}
    \begin{cases}
    \dot{S} \leq 0 & \Delta F_{mm} = \mathbf{0} ~~\text{\&}~~ F_{ext} = \mathbf{0} \quad \text{(stable)}\\
    \dot{S} \leq \Delta F_{mm}^\top\dot{x} & \Delta F_{mm} \neq  \mathbf{0} ~~\text{\&}~~ F_{ext} = \mathbf{0} \quad \text{(passive)}\\
    \dot{S} \leq (F_{ext} + \Delta F_{mm})^\top\dot{x} & \Delta F_{mm} \neq  \mathbf{0} ~~\text{\&}~~ F_{ext} \neq \mathbf{0} \quad \text{(passive)}
    \end{cases}
\end{equation}
Which means that, when there is no external force $F_{ext}=\mathbf{0}$ then, if RME provides perfect estimates then we can guarantee that the robot will track the DS $f(x)$ and converge to its target $x^*$ or trajectory $x(t)^*$. Nevertheless, if the estimate is incorrect, the system will remain passive wrt. the estimation error $\Delta F_{mm} \neq  \mathbf{0}$, which effectively reshapes the desired $f(x)$, potentially creating a spurious attractor when,
$\dot{x}^T(\Delta F_{mm}) = \dot{x}^TD(x)\dot{x}$.
Finally, when an external force exists then the system is passive wrt. the input-output port $(F_{ext} + \Delta F_{mm},\dot{x})$.

\textbf{Closed-loop behavior when $\tau_c^*$ solved as \eqref{eq:tauc_constr}}. Following the same procedure, we can write the closed-loop dynamics of the robot when the constraints are \textit{active}, while also considering the RME compensation error as follows, 
\begin{equation}
\label{eq:closed-loop-consrt-rme}
\begin{aligned}
    	M_x(q)\ddot{x} + C_x(q, \dot{q})\dot{x} + G_x(q) & = \underbrace{F_c(x)}_{\eqref{eq:passive_impedance}} - \underbrace{(J(q)^{-\top}Q^{-1} A^\top \lambda^*)}_{F_{\rm{ecbf}}} + (F_{ext} + \Delta F_{mm})\\
         M_x(q)\ddot{x} + C_x(q, \dot{q})\dot{x}  & = -D(x)\big(\dot{x}-f(x)\big) + (F_{\rm{ecbf}} + F_{ext} + \Delta F_{mm})\\
         M_x(q)\ddot{x} + \left(C_x(q, \dot{q}) + D(x)\right) \dot{x}  - \lambda_1f(x) &  =  (-F_{\rm{ecbf}} + F_{ext} + \Delta F_{mm})
\end{aligned}
\end{equation}
Note that $F_{\rm{ecbf}}\in\mathbb{R}^6$ is the pseudo-wrench the robot experiences when a kinematic constraint in joint-space is being activated; i.e., the robot will stiffen up near a constraint boundary. Following \eqref{eq:storage-cons} and \eqref{eq:power-cons}, the time-derivative of the energy storage function 
 for the closed-loop system \eqref{eq:closed-loop-consrt-rme} is, 
\begin{equation}
  \begin{aligned}
\dot{S}(x,\dot{x}) & =  \dot{x}^TM_x\ddot{x} + \frac{1}{2}\dot{x}^T\dot{M}_x\dot{x} + \lambda_1\nabla\mathcal{V}^T\dot{x}\\
& = \dot{x}^T(F_{ext} + \Delta F_{mm}) - \dot{x}^TD(x)\dot{x} - \dot{x}^T\underbrace{(J(q)^{-\top}Q^{-1} A^\top \lambda^*)}_{F_{\rm{ecbf}}}.\\
\end{aligned}  
\end{equation} 
Now we have an indefinite term in our power equation corresponding to the effect of enforcing a kinematic constraints at the joint-level by the E-CBF. Nevertheless, we can easily understand the behavior of this term. When the task-space motion of the robot $\dot{x}$ is feasible, either driven by $f(x)$ via $F_c(x)$ or an external force $F_{\rm{ext}}$ or estimation error $\Delta F_{mm}$ the E-CBF inequality $\nabla_qh(q)^\top\ddot{q} - b(q,\dot{q}) \ge 0 $ \underline{is satisfied}, then $\dot{x}^\top F_{\rm{ecbf}} \geq 0$ preserving stability and passivity as in \eqref{eq:guarantees}. 

Conversely, if the E-CBF inequality $\nabla_qh(q)^\top\ddot{q} - b(q,\dot{q}) \ge 0 $ \underline{is not satisfied} the QP will enforce it, stiffening up in the opposite direction of $\dot{x}$ and generating an opposite force rendering $\dot{x}^\top F_{\rm{ecbf}} < 0$. This is the only case when passivity is lost, yet it is a desired behavior, as we seek to enforce the constraints. This property is referred to in \cite{cbf_qp_main_paper} as \textit{passive when feasible}.
\end{proof}

\end{document}